%% file: main.tex
\documentclass[runningheads]{llncs}

\usepackage[dvipsnames,table]{xcolor}
 
\usepackage{eccv}

\input{macro}

\usepackage{algorithm}
\usepackage{algpseudocode}
\usepackage{multicol}

\usepackage{etoolbox}

\makeatletter
\expandafter\patchcmd\csname\string\algorithmic\endcsname%
      {\labelwidth 0.1em}{\labelwidth0pt\labelsep0pt}{}{}

\usepackage{eccvabbrv}

\usepackage{graphicx}
\usepackage{booktabs}

\usepackage[accsupp]{axessibility}  

\usepackage{hyperref}

\usepackage{orcidlink}

\begin{document}

\title{Semantic Residual Prompts for Continual Learning} 

\titlerunning{Semantic Residual Prompts for Continual Learning}

\author{
Martin Menabue\inst{1}\orcidlink{0000-0002-7429-9557} \and
Emanuele Frascaroli\inst{1}\orcidlink{0000-0001-7444-8826} \and
Matteo Boschini\inst{1}\orcidlink{0000-0002-2809-813X} \and
Enver Sangineto\inst{1}\orcidlink{0000-0002-5187-4133} \and
Lorenzo Bonicelli\inst{1}\orcidlink{0000-0002-9717-5602} \and
Angelo Porrello\inst{1}\orcidlink{0000-0002-9022-8484} \and
Simone Calderara\inst{1}\orcidlink{0000-0001-9056-1538}
}

\authorrunning{M. Menabue et al.}

\institute{
$^1$ AImageLab, University of Modena and Reggio Emilia, Italy \\
\email{\{name.surname\}@unimore.it}
}

\maketitle

\input{sec/0_abstract}

\input{sec/1_intro}
\input{sec/2_related}
\input{sec/3_method}
\input{sec/4_experiments}
\input{sec/5_ablations}
\input{sec/6_conclusions}

\section*{Acknowledgements}

We acknowledge the CINECA award under the ISCRA initiative, for the availability of high performance computing resources and support. This paper has been supported from Italian Ministerial grant PRIN 2020 ``LEGO.AI: LEarning the Geometry of knOwledge in AI systems'', n. 2020TA3K9N. Additionally, the research activities of Angelo Porrello have been partially supported by the Department of Engineering ``Enzo Ferrari'' through the program FAR\_2023\_DIP -- CUP E93C23000280005.

%
%
\bibliographystyle{splncs04}
\bibliography{main}

\input{suppl_arxiv}

\end{document}

%% file: macro.tex
\usepackage[dvipsnames,table]{xcolor} 

\usepackage{graphicx}
\usepackage{amssymb}
\usepackage{amsmath}
\usepackage{dsfont}
\usepackage{multirow}
\usepackage{float}
\usepackage{lipsum}
\usepackage{nicefrac}
\usepackage{circledsteps}
\usepackage{enumitem}
\usepackage{makecell}

\definecolor{cvprblue}{rgb}{0.21,0.49,0.74}
\definecolor{lightgray}{gray}{0.95}
\definecolor{midgray}{gray}{0.55}
\definecolor{steelblue}{HTML}{4D82B7}
\definecolor{greenoo}{HTML}{16A03D}
\definecolor{blueoo}{HTML}{007AB0}
\definecolor{browno}{HTML}{bc6c25}
\definecolor{paludo}{HTML}{283618}

\newcommand{\methnam}{STAR-Prompt\xspace}

\newcommand{\dpp}{DER\texttt{++}\xspace}

\newcommand{\qmarks}[1]{``{#1}''}

\usepackage{amsmath,pict2e}
\usepackage{gensymb}

\makeatletter
\newcommand{\barredsum}{%
  \DOTSB\mathop{\mathpalette\@barredsum\relax}\slimits@
}
\newcommand{\@barredsum}[2]{%
  \begingroup
  \sbox\z@{$#1\sum$}%
  \setlength{\unitlength}{\dimexpr2pt+\ht\z@+\dp\z@\relax}%
  \@barredsumthickness{#1}%
  \vphantom{\@barredsumbar}%
  \ooalign{$\m@th#1\sum$\cr\hidewidth$#1\@barredsumbar$\hidewidth\cr}%
  \endgroup
}
\newcommand{\@barredsumbar}{%
  \vcenter{\hbox{\begin{picture}(0,1)\roundcap\Line(0,0)(0,1)\end{picture}}}%
}
\newcommand{\@barredsumthickness}[1]{
  \linethickness{%
    1.25\fontdimen8
      \ifx#1\displaystyle\textfont\else
      \ifx#1\textstyle\textfont\else
      \ifx#1\scriptstyle\scriptfont\else
      \scriptscriptfont\fi\fi\fi 3
  }%
}

\makeatother

\newcommand{\vit}{ViT\xspace}
\newcommand{\backbone}{\vit-B/16\xspace}
\newcommand{\clipback}{\vit-L/14\xspace}

\newcommand{\splitcifar}{Split CIFAR-100\xspace}
\newcommand{\splitimagenet}{Split Imagenet-R\xspace}
\newcommand{\splitcars}{Split Cars-196\xspace}
\newcommand{\splitcub}{Split CUB-200\xspace}
\newcommand{\spliteurosat}{Split EuroSAT\xspace}
\newcommand{\splitresisc}{Split RESISC45\xspace}
\newcommand{\splitcropdiseases}{Split CropDiseases\xspace}
\newcommand{\splitisic}{Split ISIC\xspace}
\newcommand{\splitchestx}{Split ChestX\xspace}

\newcommand{\shortsplitcifar}{{\small{CIFAR-100}}\xspace}
\newcommand{\shortsplitimagenet}{{\small{Imagenet-R}}\xspace}
\newcommand{\shortsplitcars}{{\small{Cars-196}}\xspace}
\newcommand{\shortsplitcub}{{\small{CUB-200}}\xspace}
\newcommand{\shortspliteurosat}{{\small{EuroSAT}}\xspace}
\newcommand{\shortsplitresisc}{{\small{RESISC}}\xspace}
\newcommand{\shortsplitcropdiseases}{{\small{CropDis.}}\xspace}
\newcommand{\shortsplitisic}{{\small{ISIC}}\xspace}
\newcommand{\shortsplitchestx}{{\small{ChestX}}\xspace}

\newcommand{\clsvit}{\ensuremath{\pmb{e}_{L}^{\operatorname{CLS}}}
}
\newcommand{\result}[3]{\ensuremath{#1} \scriptsize{$\pm$\ensuremath{#2}}}
\newcommand{\resultsC}[4]{\ensuremath{#1}~\textsuperscript{(#4)}}
\newcommand{\resultC}[4]{\result{#1}{#2}{#3}{\normalsize{~\textsuperscript{(#4)}}}}
\newcommand{\resultb}[3]{\ensuremath{\mathbf{#1}} \scriptsize{$\pm$\ensuremath{#2}}}

\newcommand{\resultNV}[3]{\ensuremath{#1}}
\newcommand{\resultbNV}[3]{\ensuremath{\mathbf{#1}}}

\newcommand{\juststd}[3]{$\pm$\ensuremath{#2}}

\newcommand{\ressmall}[3]{\ensuremath{#1}{\tiny$\pm$\ensuremath{#2}}}
\newcommand{\ressmallb}[3]{\ressmall{\textbf{#1}}{#2}{#3}}

\newcommand{\faa}[1]{\ensuremath{#1}}
\newcommand{\faab}[1]{\ensuremath{\mathbf{#1}}}

\newcommand{\tit}[1]{\smallbreak\noindent\textbf{#1 }}

\usepackage{amsmath}
\DeclareMathOperator*{\argmax}{arg\,max}

%% file: sec/0_abstract.tex
\begin{abstract}
Prompt-tuning methods for Continual Learning (CL) freeze a large pre-trained model and train a few parameter vectors termed prompts. Most of these methods organize these vectors in a pool of key-value pairs and use the input image as query to retrieve the prompts (values). However, as keys are learned while tasks progress, the prompting selection strategy is itself subject to catastrophic forgetting, an issue often overlooked by existing approaches. For instance, prompts introduced to accommodate new tasks might end up interfering with previously learned prompts. To make the selection strategy more stable, we leverage a foundation model (CLIP) to select our prompts within a two-level adaptation mechanism. Specifically, the first level leverages a standard textual prompt pool for the CLIP textual encoder, leading to stable class prototypes. The second level, instead, uses these prototypes along with the query image as keys to index a second pool. The retrieved prompts serve to adapt a pre-trained ViT, granting plasticity. In doing so, we also propose a novel residual mechanism to transfer CLIP semantics to the ViT layers. Through extensive analysis on established CL benchmarks, we show that our method significantly outperforms both state-of-the-art CL approaches and the zero-shot CLIP test. Notably, our findings hold true even for datasets with a substantial domain gap w.r.t. the pre-training knowledge of the backbone model, as showcased by experiments on satellite imagery and medical datasets.
The codebase is available at \url{https://github.com/aimagelab/mammoth}.
\keywords{Continual learning \and Rehearsal-free \and Prompt-based learning}
\end{abstract}

%% file: sec/1_intro.tex
\section{Introduction}
\label{sec:intro}
\begin{figure}
    \centering
    \includegraphics[width=\textwidth]{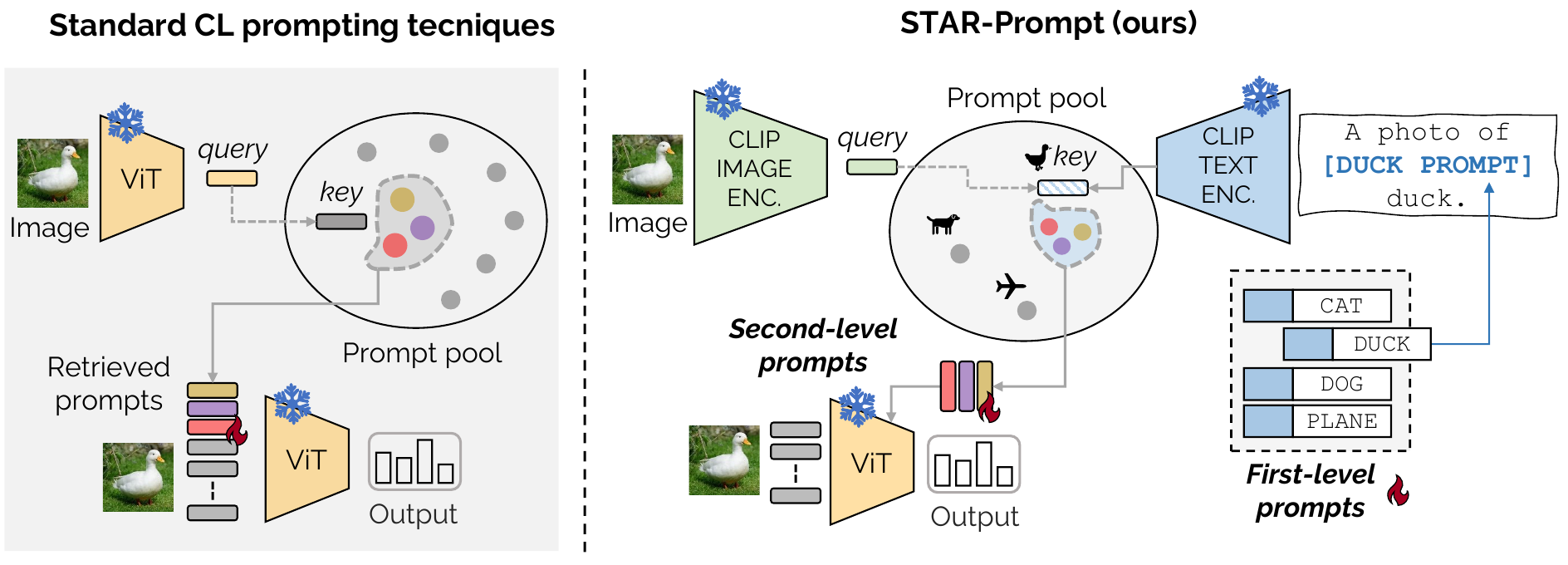}
    \caption{Comparison between current approaches (left) and our \methnam (right), regarding their prompting selection strategies. To enhance the stability of the pool selection, we exploit the multi-modal CLIP embedding space and compute a similarity between the image and prompt-learned class-prototype keys. Afterwards, the retrieved values are used as prompts for another backbone, \ie an ImageNet pre-trained ViT.
    }
    \label{fig:introfigure}
\end{figure}
While human beings can easily acquire new knowledge and remember past information, several studies~\cite{ratcliff1990connectionist,mccloskey1989catastrophic} have shown that Artificial Intelligence models struggle to replicate this behavior. Specifically, neural networks tend to forget previously solved tasks while learning new ones, an issue known as \textit{catastrophic forgetting}~\cite{mccloskey1989catastrophic}. Continual Learning (CL)~\cite{de2021continual, parisi2019continual} models a scenario in which data from old tasks are no longer available when training on new tasks, and the model needs to trade-off the preservation of past information (\textit{stability}) with the adaptation to the new data (\textit{plasticity}).

The emergence of large-scale pre-trained Transformers~\cite{vaswani2017attention} and foundation models~\cite{DBLP:journals/corr/abs-2108-07258} has recently changed the adaptation paradigm, usually moving towards \textit{parameter-efficient} plasticity, in which only a few parameters are optimized to solve a new task while keeping most of the model frozen. A successful example is \textit{prompt tuning}, in which a few learnable vectors (\qmarks{soft prompts}) are appended to the input image tokens. For instance, CoOp~\cite{zhou2022learning} learns a context prompt, which is concatenated with the textual name of the class label and fed to the textual encoder of CLIP~\cite{radford2021learning}. The latter generates a {\em class prototype} in feature space, which is used to classify the images of the task at hand. Similarly, Visual Prompt Tuning (VPT)~\cite{DBLP:conf/eccv/JiaTCCBHL22} learns task-specific visual prompts for each layer of an ImageNet pre-trained Vision Transformer (ViT)~\cite{dosovitskiy2020image}, while CoCoOp~\cite{zhou2022conditional} extends CoOp by conditioning its prompts on the input image.

Prompt tuning has recently been adopted in different CL proposals~\cite{wang2022l2p,wang2022dualprompt,smith2023coda,Wang_2023_CVPR,chen2023promptfusion} with good results. These methods, inspired by CoOp, CoCoOp, and VPT, usually devise a shared pool of key-value pairs to manage the prompts learned during each round. As shown in Fig.~\ref{fig:introfigure} (left), the input image is used as a query to retrieve a subset of \textit{relevant} prompts from the pool. To make the selection strategy effective, the keys are learned along with the values (\ie, the prompts). The learning criteria for these keys usually involve a form of weak supervision that pulls selected keys closer to corresponding query features, with further orthogonality constraints to encourage diversification~\cite{smith2023coda}.

Despite the extensive application in incremental scenarios, prompting approaches fall into a pitfall, which is the main focus of our work. As the key space is continuously updated with no further access to past queries/tasks, the selection mechanism is itself subject to \textit{catastrophic forgetting} and possible misalignments between queries and keys. As outlined in Sec.~\ref{sec:discussion}, this causes interference in the prompt selection (\ie, prompts of a certain task are mapped to unrelated tasks), thus leading to sub-optimal performance.

We hence propose to employ a foundation model like CLIP to realize the query-key matching mechanism, yielding a more stable selection strategy. In fact, it has been shown that the fine-tuning regime of CLIP models favors stability over plasticity~\cite{chen2023promptfusion}, as distinct parameters can be naturally devoted to distinct concepts, mitigating their interference. However, to enhance plasticity, we use the prompts retrieved through CLIP to condition another backbone, specifically an ImageNet pre-trained ViT. Notably, this leads to a \textbf{two-level prompting pipeline} (Fig.~\ref{fig:introfigure} right), which is the main technical contribution of our work. Briefly: as in CoOp and AttriCLIP~\cite{wang2023attriclip}, we train class-specific prompts (\textbf{first-level prompt pool}) to be fed to the CLIP textual encoder. Conversely, we exploit the resulting embedding not for computing classification scores, but as keys of a \textbf{second} prompt pool. The retrieved \textbf{second-level prompts} are used to adapt the ImageNet pre-trained ViT, which provides the final output prediction.

Our secondary contribution involves a novel mechanism to prompt the pre-trained ViT, utilizing an \textit{additive residual} technique to adapt it. While most methods~\cite{wang2022dualprompt,smith2023coda} \textit{concatenate} the learned prompts to the arguments of the Multi-head Self-Attention (MSA) layer (see \cref{sec:related}), our second-level prompts are used as \textit{semantic residuals} which are {\em added} before the MLP layer input. While there is no guarantee that a pre-trained and frozen MSA layer will take into account all the prompt tokens, our additive mechanism forces the ViT embeddings to include CLIP-derived semantics.

Finally, inspired by SLCA~\cite{zhang2023slca}, where first- and second-order feature statistics are used to generate synthetic samples of past tasks, we adopt a generative replay method in both stages of our approach. However, differently from SLCA, we use a Mixture of Gaussians (MoGs) to model the \textit{multimodal distribution} of each class, leading to a more effective generation process.

To assess our intuitions, we test our approach termed \textbf{\methnam} (Semantic Two-level Additive Residual Prompt) on \textbf{nine} image classification datasets. These encompass well-established CL benchmarks, such as \splitimagenet and \splitcifar~\cite{wang2022dualprompt, smith2023coda, zhang2023slca, chen2023promptfusion}, but also satellite and medical settings, hence covering domains that \textbf{differ significantly}~\cite{oh2022understanding, cui2018domaintransfer, radford2021learning} from the pre-trainings of both ImageNet and CLIP. As our experiments outline, prompting-based CL approaches struggle with such large domain gaps, often being outperformed by older methods relying on the good old rehearsal schema.
The results show a remarkable $\texttt{+}5.96$ average improvement in final accuracy \wrt the state of the art, with peaks of $\texttt{+}3$ on \splitimagenet~and $\texttt{+}11$ on \splitcars. Importantly, we prove the resilience of our framework to severe domain shifts. For example, when the zero-shot CLIP model struggles (\eg, achieving only \faa{54.50\%} accuracy on \spliteurosat), our two-level approach instead excels (\ie, \faa{93.70\%}), underscoring the merits of a stable and continuous adaptation over the simple reuse of frozen large pre-trained models.
We therefore acknowledge the following \textbf{contributions}:
\begin{itemize}[noitemsep]
    \item We shed light on the stability issue of prompt selection proposing a two-level prompting strategy that builds upon the stability of foundation models.
    \item We present a novel prompting approach based on semantic residuals.
    \item We extend the SLCA feature generation to multi-modal distributions. 
    \item We deepen the current understanding of incremental adaption, underscoring that large pre-trained models still require adaptability.
\end{itemize}

%% file: sec/2_related.tex
\section{Related work}
\label{sec:related}
\tit{Continual Learning.} CL approaches are usually classified according to the strategy they adopt against forgetting. \textbf{\textit{Regularization-based methods}} adjust the loss function to restrict significant changes in either the model weights or its activations when moving to other tasks~\cite{kirkpatrick2017overcoming,li2017learning,ritter2018online,mirzadeh2020understanding}. \textbf{\textit{Architectural methods}} alter the structure of the model by devoting separate groups of parameters to different tasks~\cite{rusu2016progressive,mallya2018packnet,serra2018overcoming}. Finally, \textbf{\textit{Rehearsal-based methods}} alleviate forgetting by storing a subset of samples from past tasks in a memory buffer, which are then optimized along with the data of the current task~\cite{ratcliff1990connectionist,rebuffi2017icarl,hou2019learning,buzzega2020dark,caccia2022new}. A sub-category of rehearsal methods is based on \textbf{generative replay}, where real samples are replaced by synthetically generated ones~\cite{robins1995catastrophic,shin2017continual,kemker2017fearnet}.

\tit{Prompting methods.} The first attempt to use prompt tuning in CL is Learning to Prompt (\textbf{L2P})~\cite{wang2022l2p}, where a pool of prompts is shared by all tasks, and the input image is used as the query to select the most appropriate prompts from this pool. \textbf{DualPrompt}~\cite{wang2022dualprompt} introduces a hierarchy of general and task-specific prompts and adopts \textbf{prefix-tuning} over prompt tuning, where the former prepends prompts to the keys and the values of MSA layers rather than prepending prompts to the input tokens. Prefix-tuning is also adopted in \textbf{CODA-Prompt}~\cite{smith2023coda}, which additionally introduces an end-to-end prompting mechanism and an orthogonality soft-constraint to encourage prompt independence (both of which we adopt, see \cref{sec:details}).

Recently, several works have investigated the application of CLIP~\cite{radford2021learning} in incremental settings. In \textbf{AttriCLIP}~\cite{Wang_2023_CVPR} the prompts are fed to the CLIP's textual encoder, while \textbf{PromptFusion}~\cite{chen2023promptfusion} takes inspiration from VPT and trains prompts for the textual encoder jointly with visual prompts for a pre-trained ViT. Differently from PromptFusion, which uses these prompts in \textit{parallel} fusing the corresponding classification scores, we instead employ them \textit{sequentially}. Specifically, the textual prompts are used to produce class-specific keys to index a second pool of prompts.

%% file: sec/3_method.tex
\begin{figure*}[t]
 \centering
 \includegraphics[width=\textwidth]{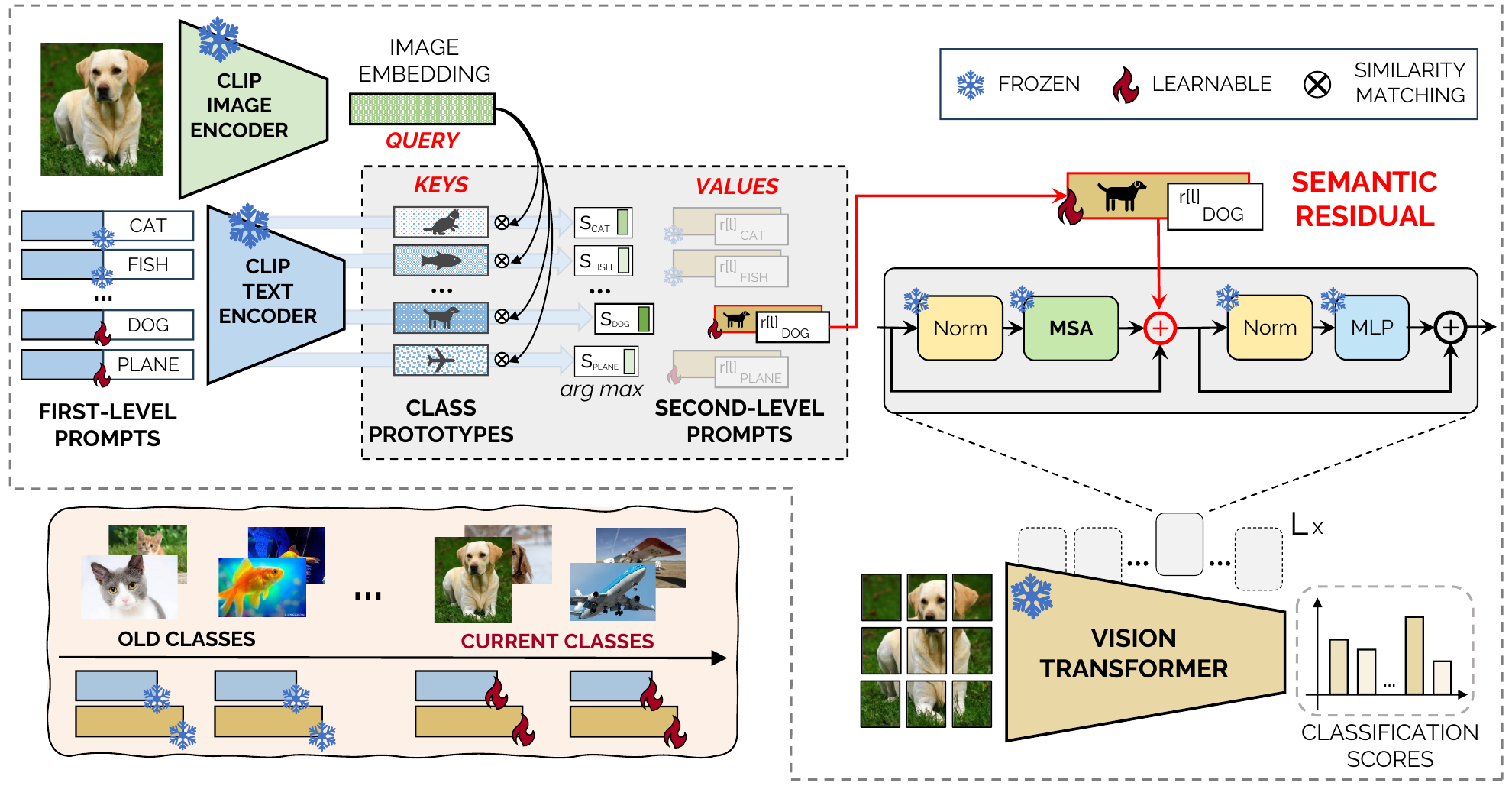}
 \caption{The architecture of \textbf{\methnam}. The bottom left box illustrates our CL setting, in which first- and second-level prompts of the old tasks are frozen.}
 \label{fig:framework}
\end{figure*}
\section{Method}
\label{sec:method}
\tit{Problem setting.} In Continual Learning, a model is trained on a sequence of $T$ tasks, where each task $D_t$, $t=1,\dots,T$ is composed of sample-label pairs $(x_i,y_i)$, $i=1, \dots, |D_t|$. In this paper, we adopt a class-incremental setting~\cite{van2022three}, in which the set of classes ${\cal Y}_t$ of task $D_t$ is disjoint from ${\cal Y}_{t'}$ (with $t \neq t'$), and the task identity $t$ is unknown at inference time. For simplicity, we assume that $|{\cal Y}_1| = \dots = |{\cal Y}_T|= N$. Importantly, when the model is trained on $D_t$, the training data from the old tasks $D_{1:t-1}$ are inaccessible.
\subsection{Overview}
\label{sec:setting}
Our method, shown in~\cref{fig:framework}, relies on two pre-trained architectures:
\begin{itemize}
    \item the CLIP model~\cite{radford2021learning}, \ie the image $E_{vis}(\cdot)$ and the text $E_{txt}(\cdot)$ encoders. 
    \item an ImageNet pre-trained Vision Transformer~\cite{dosovitskiy2020image} $f(\cdot)$.
\end{itemize}
In a nutshell, every parameter of these networks is frozen during training. To adapt them in a parameter-efficient manner while minimizing interference, we devise a \textbf{two-level codebook strategy}. The first one provides learnable context passed as input to the CLIP text encoder: by doing so, we produce \textit{stable} class prototypes. The second level leverages these text-level prototypes as keys of a second codebook of prompts, which serve to provide a more \textit{plastic} behaviour to the second model $f(\cdot)$. 

Specifically, the first-level prompting stage resembles CoOp and conditions the text encoder $E_{txt}(\cdot)$ by concatenating a class-specific learnable vector to the input sequence. However, differently from CoOp, we do not directly exploit the learned textual class prototypes to perform classification. In fact, we use them as \textit{keys} for comparison against the visual \textit{query}, derived by applying the CLIP visual encoder $E_{vis}(\cdot)$ on the input image. Based on the resulting similarities, we retrieve a second-level learnable prompt $R$, termed \textbf{semantic residual vector}, aiming to transfer class-specific semantics to the ViT $f(\cdot)$.

In particular, we devise a specific residual vector for each layer of $f(\cdot)$. Formally, indicating with $\pmb{e}^l$ the output of the $l$-th block of $f(\cdot)$ containing MSA, MLP, and LayerNorm (LN) layers~\cite{dosovitskiy2020image}, 
\begin{align}
 \pmb{e}' &= \operatorname{MSA}(\operatorname{LN}(\pmb{e}_l)) + \pmb{e}_l \label{eq.ViT-MSA} \\
 \pmb{e}_{l+1} &= \operatorname{MLP}(\operatorname{LN}(\pmb{e}')) + \pmb{e}', \label{eq.ViT-MLP}
\end{align}
we exploit the first residual connection to inject the semantic residual vector $R$:
\begin{equation}
\pmb{e}' = \operatorname{MSA}(\operatorname{LN}(\pmb{e}_l)) + \pmb{e}_l + R.
\end{equation}
Ultimately, the classification scores are derived from the final classification layer of $f(\cdot)$. We provide the details in the following subsections.
\subsection{\methnam: Two-Level Prompt Tuning}
\label{sec:two-level-prompting}
In the first prompting stage, we associate a learnable prompt $p_c$ to each class $y_c \in {\cal Y}_t$. The prompt $p_c$ consists of a single learnable token that is concatenated with the initial word embedding of the class $y_c$, indicated with $\operatorname{[CLS-NAME]}$:
\begin{equation}
 \label{eq.class-prototypes}
\pmb{w}_c = E_{txt}([p_c; \operatorname{[CLS-NAME]}]).
\end{equation}
For instance, if $y_c$ refers to the class `\textit{dog}', then $\operatorname{[CLS-NAME]}$ is the embedding of the word `\textit{dog}'. These two embeddings are fed to $E_{txt}(\cdot)$ to get a final textual representation $\pmb{w}_c \in \mathbb{R}^d$.

While CoOp utilizes $\pmb{w}_c$ as a class prototype to classify the image $x$, we treat $\pmb{w}_c$ as a \textbf{class-specific key}. Specifically, we use these class-specific keys to retrieve a second set of learnable prompts $Q_c \in \mathbb{R}^{L \times d'}$. These prompts aim to adapt each layer of the ImageNet pre-trained Vision Transformer $f(\cdot)$. In particular, let $l<L$ denote the $l$-th layer of $f(\cdot)$ and $d'$ its embedding dimension. Then, $Q_c[l] \in \mathbb{R}^{d'}$ represents the second-level class-specific prompts used to adapt the $l$-th layer of $f(\cdot)$.

In detail, the second-level prompting stage leverages the CLIP multi-modal embedding space. Given the visual embedding of the input image $\pmb{z} = E_{vis}(x)$ (query), we compute its cosine similarity $\operatorname{sim}_c = \langle \pmb{z}, \pmb{w}_c \rangle$ with each class prototype key $\pmb{w}_1, \dots, \pmb{w}_{Nt}$. Each of these keys is derived from the first-level prompting stage and corresponds to one of the $Nt$ classes observed so far.

Afterwards, we build the \textbf{semantic residual} $\mathbf{R} \in \mathbb{R}^{L \times d'}$ by using the prompt $Q_{c_k}$ of the class with the \textbf{highest} cosine similarity, as follow:
\begin{align}
\mathbf{R} &= \operatorname{sim}_{c_k} \cdot \ Q_{c_k}, \; \label{eq.semantic-residual} \\
\textrm{where} \quad c_k &= \argmax_{c = 1, \dots, Nt} \operatorname{sim}_{c}. \label{eq.training-time-selection}
\end{align}
In \cref{eq.semantic-residual}, the term $\operatorname{sim}_{c_k}$ weighs the residuals with the confidence CLIP assigns to the selected key $\pmb{w}_{c_k}$. As shown in \cref{sec:ablations}, such a modulating operation yields indeed a beneficial empirical effect. 

The residuals $\mathbf{R}$ are then used to transfer CLIP-based class-specific cues to each MLP layer of $f(x)$. We hence replace \cref{eq.ViT-MSA} with:
\begin{equation}
\label{eq.residual-in-ViT-MSA} 
 \pmb{e}' = \operatorname{MSA}(\operatorname{LN}(\pmb{e}_l)) + \pmb{e}_l + \mathbf{R}[l].
\end{equation}
Please note that $\mathbf{R}[l]$ is a single $d'$-dimensional vector that applies uniformly across the entire sequence of visual tokens. In other words, to ensure dimensional compatibility for summation, $\mathbf{R}[l]$ is repeated for each token in the sequence.
\tit{Main training objective.} For each task $D_t$, we extend the two codebooks adding the class-specific prompts of the current task. In detail, the set of first-level prompts is computed as: ${\cal P}_t := {\cal P}_{t-1} \cup \{ p_c~|~c \in {\cal Y}_t \}$, while the second-level codebook is obtained by: ${\cal Q}_t := {\cal Q}_{t-1} \cup \{ Q_c~|~c \in {\cal Y}_t \}$. To avoid forgetting, we \textbf{freeze} the prompts of the old tasks (${\cal P}_{t-1}$ and ${\cal Q}_{t-1}$) and train only those corresponding to the current task. Although both levels may be trained jointly and end-to-end, we strive for simplicity and structure the optimization process as a two-stage training procedure. Specifically, for both codebooks, we use the cross-entropy as the main loss function:
\begin{equation}
\label{eq.common-cross-entropy}
{\cal L}_{\operatorname{CE}} = - \frac{1}{|D_t|} \sum_{i=1}^{|D_t|} \log p(y_i = c|x_i),
\end{equation}
\noindent where, in the first stage, the posterior probabilities are computed as:
\begin{equation}
\label{eq.1-stage-posterior}
 p(y_i = c|x_i)= \frac{\operatorname{exp}({\langle \pmb{z}_i, \pmb{w}_c \rangle}/\tau)}
 {\sum_{c' \in {\cal Y}_t} \operatorname{exp}({\langle \pmb{z}_i, \pmb{w}_{c'} \rangle}/\tau)}, 
\end{equation}
\noindent with $\tau$ indicating the temperature parameter of CLIP. Conversely, in the second stage, the posteriors in~\cref{eq.common-cross-entropy} consider the $\operatorname{CLS}$ token $\clsvit = f(x_i;\mathbf{R})$ of the last layer $L$ of $f(\cdot)$~\cite{dosovitskiy2020image}, as follows:
\begin{equation}
\label{eq.2-stage-posterior}
 p(y_i = c|x_i)= g_{\theta_t} (\clsvit),
\end{equation}
\noindent where $g_{\theta_t} (\cdot)$ represents the classification head of the current task $t$. This module comprises of a linear projection followed by the softmax layer. The parameters $\theta_t$ of the projection layer are initialized at the beginning of task $t$, and trained while keeping the classification heads of the old tasks frozen.
\subsection{Additional Details}
\label{sec:details}
\tit{Query weighting.} Similarly to CODA-Prompt~\cite{smith2023coda}, we train class-specific weight vectors $A_c \in \mathbb{R}^d$ to weigh the importance of the visual features with respect to the $c$-th class. Specifically, each $\pmb{z} = E_{vis}(x)$ is element-wise multiplied with $A_c$, and in~\cref{eq.semantic-residual} we replace $\operatorname{sim}_c = \langle \pmb{z}, \pmb{w}_c \rangle$ with $\langle \pmb{z} \odot A_c , \pmb{w}_c \rangle$.

\tit{Orthogonality.} We also follow CODA-Prompt and discourage possible correlations between old and new prompts. This is done by minimizing the following two loss functions for the two codebooks respectively:
\begin{gather}
{\cal L}_{\operatorname{OP}} = \sum_{c \in {\cal Y}_t, c' \in \operatorname{past}(t)} \langle \pmb{p}_{c'}, \pmb{p}_c \rangle,
\label{eq.orthog-prompt-1} \\
{\cal L}_{\operatorname{OQ}} = \frac{1}{L} \sum_{l=1}^L \sum_{c \in {\cal Y}_t, c' \in \operatorname{past}(t)} \langle Q_{c'}[l], Q_c[l] \rangle.
\label{eq.orthog-prompt-2} 
\end{gather}
where $\operatorname{past}(t) = \{c' \lvert c' \in {\cal Y}_{t'}, t'<t\}$ indicates the set of past classes.
\subsection{Discussion and comparisons to related works}
\label{sec:discussion}
Our main contribution relies on the use of a foundation model to enhance the \textbf{stability} of the prompt selection. Indeed, we regard a prompting strategy as \textit{\textbf{stable}} if, given a task, the query-key selection strategy retrieves the set of prompts \textbf{relevant} to that task, and not those introduced for subsequent tasks. In the following, we dissect why existing approaches overlook stability and how we exploit CLIP to enhance it.

\tit{On the stability of prompting-based CL methods.} Like our method, existing approaches such as L2P, DualPrompt, and CODA-Prompt employ a query-key matching mechanism to fetch the prompts utilized for adapting the pre-trained ViT. Notably, they learn new keys \textbf{from scratch} when a new task is presented, and define a customized loss term to incentivize the learned keys to be aligned with the visual queries of the current task (extracted using the pre-trained ViT itself). However, these learned keys often \textbf{lack} a clear semantic interpretation, and could be theoretically used to represent distant concepts.
\input{figures/retrieval_acc}

Concerning this, we argue that learning keys as such is susceptible to interference between tasks. Indeed, since keys are continuously adjusted to align with the queries of the current task, they may drift and no longer match with the queries of earlier tasks. Or, when introducing novel keys, they could interfere with those already present in the pool, and be selected to classify examples of previous tasks. We assess such an intuition by analyzing the prompts retrieved by different methods -- \ie, DualPrompt, CODA-Prompt, and our approach \methnam -- on the \splitimagenet dataset. In detail, Fig.~\ref{fig:retrieval_interference_left}, shows the proportion of testing samples of the first task $D_1$ retrieving the \textit{right} prompts -- namely, those corresponding to prompts learned during the first task $D_1$. As can be seen, the precision of the selection strategies of both DualPrompt and CODA-Prompt \textbf{decreases} as the tasks progress, indicating that more and more examples of $D_1$ use prompts of subsequent tasks. Our approach is instead much more stable, as also evident in Fig.~\ref{fig:retrieval_interference_right}, which depicts the full confusion matrices computed after the final task. While the confusion matrix of \methnam is almost diagonal, the prompt selection of DualPrompt and CODA-Prompt yields significant task confusion. As outlined in the suppl. materials, we verified these outcomes also on \textbf{other domains} (\ie, \textbf{in aerial and medical datasets}).

The reason at the base of the stability of our \methnam is twofold. On the one hand, we do not start from scratch when learning the keys, but we adapt the CLIP text encoder, which allows us to devote \textit{distinct learnable parameters} to \textit{distinct classes}. Such an approach promotes a greater \textbf{modularity}, and has been already acknowledged as more stable by the authors of PromptFusion~\cite{chen2023promptfusion}. Differently from them, however, we exploit the stability of the text encoder not for direct prediction, but to fetch a second codebook of prompts.

On the other hand, we provide \textbf{explicit supervision} (\cref{eq.common-cross-entropy}) to enforce class-driven \textbf{separation} in the key space. This differs from other CL prompting approaches, which settle to \textit{weaker} constraints (\eg, relying on orthogonality or aligning the query with the closest key in the pool). Hence, since $\pmb{w}_c$ keys are both stable and separated, the second-level prompts $Q_c$ can learn class-specific information, {\em reducing the interference among classes/tasks}.

\tit{Prompting mechanism.} In most methods, prompts are \textbf{prepended} with the input sequence \textbf{before} each MSA layer. Another common solution is \textbf{prefix tuning} (\cref{sec:related}), which intervenes only on the keys and the values of the MSA layer. Since the frozen MSA layer weighs the importance of each prompt, we argue that there is no barrier preventing some of these tokens from being disregarded. Conversely, our semantic residuals are \textbf{added} after the MSA layer (see \cref{eq.residual-in-ViT-MSA}). This is similar to the shifting parameters of FiLM layers~\cite{perez2018film} used to adapt the Batch Norm~\cite{DBLP:conf/icml/IoffeS15} of CNNs. However, we do not have a parameter \qmarks{generator} network~\cite{perez2018film}, and the residuals for the Vision Transformer MLPs are computed using CLIP-guided shifting vectors.

\tit{Computational cost.} As detailed in suppl. materials, the computational demand associated with \methnam is comparable with that of most prompt-based approaches (\eg, L2P, CODA-Prompt, etc.), requiring two forward passes. The key difference is that these methods rely on the same backbone, whereas we employ two backbones with distinct parameters.
\subsection{Multi-Modal Generative Replay}
\label{subsec:mogs}
\input{algo/main_algo}
In SLCA, class-specific features are used to estimate the statistics of a Gaussian representing the class distribution. Then, subsequent tasks use this Gaussian to generate synthetic features and perform generative replay. To capture multiple peaks in the distribution of each class, we extend this idea by introducing, in both stages of our training, a Mixture of Gaussians (MoGs) representation of the feature distribution. Specifically, in the first stage, for each class $c$, we fit a MoGs $\mathcal{G}_{c}$ on the CLIP visual features $\pmb{z}_i = E_{vis}(x_i) \ \forall \ (x_i, y_i=c) \in D_t$:
\begin{equation}
\label{eq.MoG}
\mathcal{G}_{c} = \mathcal{G}(\cdot; \pmb{\phi_c}) = \sum _{m=1}^{M}\phi^{c}_{m}{\mathcal {N}}(\cdot; {{{\mu}^{c} _{m},{\Sigma}^{c}_{m}}}). 
\end{equation}
The parameter vector $\pmb{\phi_c}$ contains the $M$ means, covariances and corresponding Gaussian component weights, and it is estimated by Expectation-Maximization (EM). $\mathcal{G}_{c}$ is then used in the subsequent tasks $t'$ ($t' \geq t$) to generate $n = 256$ synthetic features $\tilde{\pmb{z}}$ of class $c$. Using this generative replay technique, we obtain:
\begin{equation}
\label{eq.cross-entropy-reharsal}
{\cal L}_{\operatorname{GR}}^{P} = - \frac{1}{n Nt}
 \sum_{c=1}^{Nt} \sum_{i=1}^n \log p(y_i = c|\tilde{\pmb{z}}_i),
\end{equation}
\noindent where $p(y_i = c|\tilde{\pmb{z}}_i)$ is the posterior probability for a synthetic visual point $\tilde{\pmb{z}}_i \sim \mathcal{G}_{c}$:
\begin{equation}
\label{eq.1-stage-posterior-reharsal}
 p(y_i = c|\tilde{\pmb{z}}_i)= \\
 \frac{\operatorname{exp}({\langle \tilde{\pmb{z}}_i, \pmb{w}_c \rangle}/\tau)}
 {\sum_{c'=1}^{Nt} \operatorname{exp}({\langle \tilde{\pmb{z}}_i, \pmb{w}_{c'} \rangle}/\tau)}.
\end{equation}
\noindent Note that, in the denominator, we use all the $Nt$ classes observed so far.

Similarly, in the second training stage, we use the ViT features $\clsvit = f(x_i; \mathbf{R})$ of the current task to estimate the parameters of a second MoGs, indicated with $\mathcal{H}_{c}$, as for \cref{eq.MoG}. In the rehearsal phase, we replace $\clsvit$ in~\cref{eq.2-stage-posterior} with $n$ synthetic features sampled from each $\mathcal{H}_c$, and, analogously to~\cref{eq.1-stage-posterior-reharsal,eq.cross-entropy-reharsal}, we compute ${\cal L}_{\operatorname{GR}}^{Q}$ (details in the suppl. materials).

The whole training algorithm is shown in~\cref{alg.1}. For ease of presentation, we do not split the dataset in mini-batches when computing a step of gradient descent. The algorithm shows that, in the first training stage, only the first-level prompts are updated (\ie, $p_c, c \in {\cal Y}_t$). Conversely, in the second stage, we train the second-level prompts $Q_c$, the query weighting parameters $A_c$, and the linear layer of $g_{{\theta}_t}(\cdot)$. Specifically, in the second-stage rehearsal phase, we update the heads of all the tasks observed so far (\ie, $\theta_{t' \le t}$).

%% file: figures/retrieval_acc.tex
\begin{figure*}[t]
    \begin{minipage}[t]{.39\textwidth}
        \centering
        \begin{tabular}{c}
        \includegraphics[trim={0 .8cm 0 .8cm},clip,width=\textwidth]{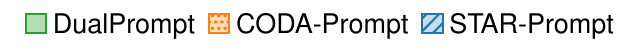} \\
        \includegraphics[width=\textwidth]{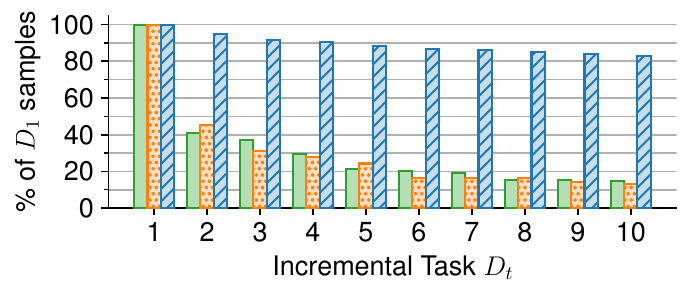} \\
        \end{tabular}
        \subcaption{}\label{fig:retrieval_interference_left}
    \end{minipage}
    \hfill
    \begin{minipage}[t]{.59\textwidth}
        \centering
        \setlength{\tabcolsep}{2pt}
        \begin{tabular}{ccc}
        \includegraphics[width=0.30\textwidth]{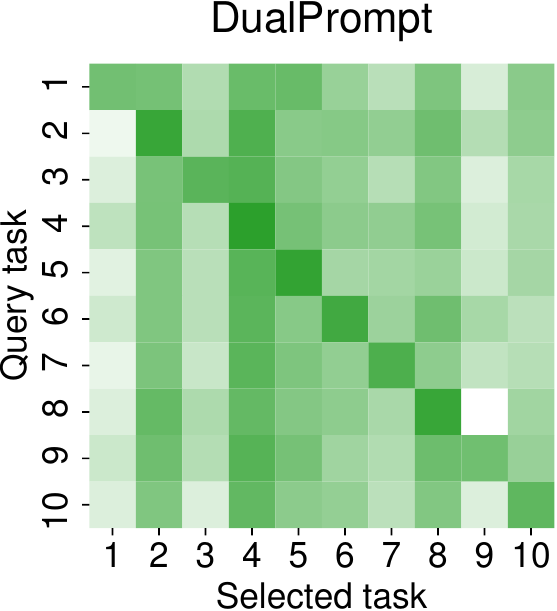} &
        \includegraphics[width=0.30\textwidth]{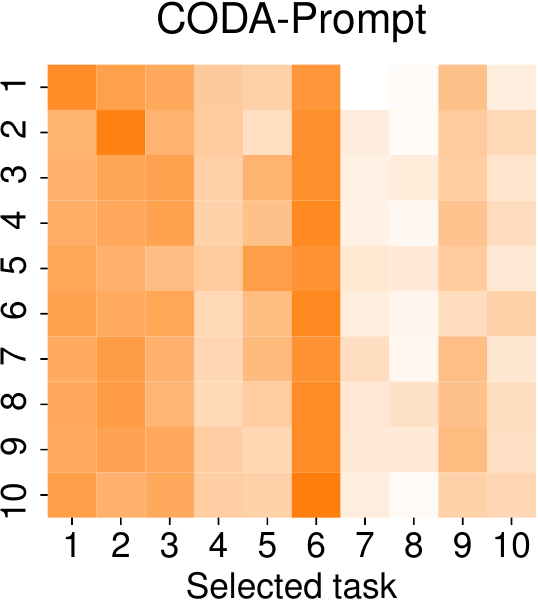} & 
        \includegraphics[width=0.30\textwidth]{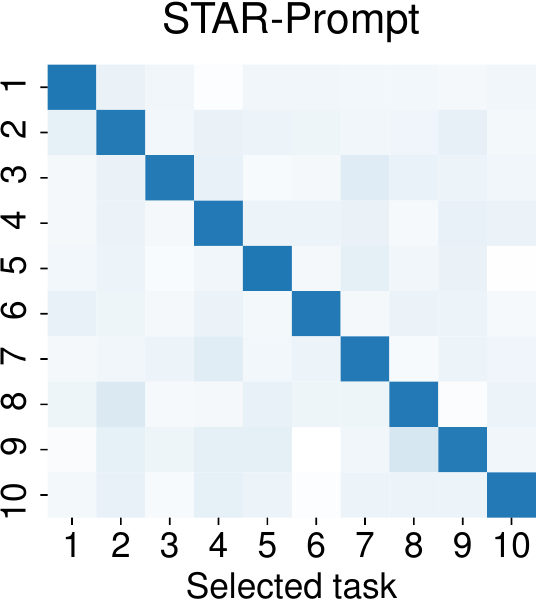} \\
        \end{tabular}
    \subcaption{}\label{fig:retrieval_interference_right}
    \end{minipage}  
    \label{fig:1-2}
    \vspace{-1em}
    \caption{Prompt retrieval on \splitimagenet. \textit{Left:} for the test set of the first task, the portion of examples using prompts of the correct task $D_1$ during the incremental training. \textit{Right:} we extend the analysis to all the tasks, computed at the end of the incremental training. In each confusion matrix, the $y$ axis represents the task of the query sample, while the $x$ axis shows the task of the corresponding selected key.}
    \vspace{-1em}
\end{figure*}

%% file: algo/main_algo.tex
\setlength\fboxsep{0pt}
\renewcommand\algorithmicindent{.7em}
\algrenewcommand\alglinenumber[1]{\tiny #1:}
\begin{algorithm}[t]
 \caption{Training of \methnam on the current task}
 \label{alg.1}
\begin{algorithmic}[1]

\Require Index of the current task $D_t, \ t \in {1,\dots,T}$, the orthogonality loss weighting coefficient $\lambda$, the number of $M$ Gaussian components, the number of $E_1$ and $E_2$ training epochs with real and generated samples (respectively), learning rate $lr$.
\begin{multicols}{2}
\Statex \textcolor{blue}{\textbf{First stage}} $\rightarrow$ \textit{goal}: learning first-level prompts $p_c \vert_{c \in \{ 1, ..., N\}}$ for each class $c \in {\cal Y}_t$ of the current task $t$.
\smallskip
\For{$ep := 1, \dots, E_1$}{}
    \State ${\cal L}_{\operatorname{CE}} \xleftarrow{}$ use \cref{eq.1-stage-posterior,eq.common-cross-entropy}
    \State ${\cal L}_{\operatorname{OP}} \xleftarrow{}$ use \cref{eq.orthog-prompt-1}
    \State $p_c \xleftarrow{} p_c - lr \cdot \nabla_{p_c \vert_{c \in {\cal Y}_t} } {\cal L}_{\operatorname{CE}} + \lambda {\cal L}_{\operatorname{OP}}$
\EndFor
\State \# \textit{apply Generative Replay (\cref{subsec:mogs})}
\State $\mathcal{G}_{c} \xleftarrow{} \operatorname{EM}(\{E_{vis}(x)\}_{x, y = c \in D_t}) \ \forall c \in {\cal Y}_t$
\smallskip
\For{$ep := 1, \dots, E_2$}{}
    \State \# \textit{create samples from} $\mathcal{G}_{c'} \vert_{c' \in {\cal Y}_1, \dots, {\cal Y}_t}$
    \State $\mathcal{L}_{\operatorname{GR}}^{P} \xleftarrow{}$ use \cref{eq.cross-entropy-reharsal}
    \State $p_c \xleftarrow{} p_c - lr \cdot \nabla_{p_c \vert_{c \in {\cal Y}_t}} {\cal L}_{\operatorname{GR}}^{P}$
\EndFor
    \columnbreak
\Statex \textcolor{blue}{\textbf{Second stage}} $\rightarrow$ \textit{goal}: learning second-level prompts $Q_c$, query weights $A_c$ $\forall c \in {\cal Y}_t$, and the classifiers $\theta_{t'} \vert_{t' \in \{ 1, ..., t \}}$.
\smallskip
\State \texttt{param} $\xleftarrow{} {\{Q_c, A_c\}}_{c \in {\cal Y}_t} \cup \theta_t$
\For{$ep := 1, \dots, E_1$}{}
    \State $\mathcal{L}_{\operatorname{CE}} \xleftarrow{}$ use \cref{eq.2-stage-posterior,eq.common-cross-entropy}
    \State $\mathcal{L}_{\operatorname{OQ}} \xleftarrow{}$ use \cref{eq.orthog-prompt-2}
    \State $\texttt{param} \xleftarrow{} \texttt{param} - lr \cdot \nabla_{\texttt{param}} \ {\cal L}_{\operatorname{CE}} + \lambda {\cal L}_{\operatorname{OQ}}$
\EndFor
\smallskip
\State \# \textit{apply Generative Replay (\cref{subsec:mogs})}
\State $\mathcal{H}_{c} \xleftarrow{} \operatorname{EM}(\{f(x_i; \mathbf{R})\}_{x, y = c \in D_t}) \ \forall c \in {\cal Y}_t$
\smallskip
\For{$ep := 1, \dots, E_2$}{}
    \State \# \textit{create samples from} $\mathcal{H}_{c'} \vert_{c' \in {\cal Y}_1, \dots, {\cal Y}_t}$
    \State $\theta_{t'} \xleftarrow{} \theta_{t'} - lr \cdot \nabla_{\theta_{t'}} {\cal L}_{\operatorname{GR}}^{Q} \ \forall \ t' \in \{ 1, ..., t \}$
\EndFor
\end{multicols}
\end{algorithmic}
\end{algorithm}

%% file: sec/4_experiments.tex
\section{Experiments}
\label{sec:experiments}
We evaluate our approach on varying benchmarks. We follow the current literature dealing with pre-trained CL models~\cite{wang2022dualprompt,smith2023coda,zhang2023slca,chen2023promptfusion} and cover conventional image datasets such as \shortsplitcifar and \shortsplitimagenet. Then, we assess the adaptability of these pre-trained methods by moving to settings with decreasing domain similarity w.r.t. the Image-Net pre-training~\cite{oh2022understanding, cui2018domaintransfer}. In particular, we test on the following domains (more details can be found in suppl. materials):
\begin{itemize}[noitemsep]
    \item \textbf{Natural domain}: \textit{\splitcifar}~\cite{krizhevsky2009cifar} and \textit{\splitimagenet}~\cite{hendrycks2021many}, with respectively $100$ and $200$ classes split into $10$ tasks.
    \item \textbf{Specialized domain}: following~\cite{zhang2023slca}, we adopt \textit{\splitcars}~\cite{krause20133d} and \textit{\splitcub}~\cite{wah2011cub} as fine-grained scenarios. Classes are split across $10$ tasks.
    \item \textbf{Aerial domain}: datasets comprising RGB satellite images for land use and land cover classification. We use \textit{\spliteurosat}~\cite{helber2018introducing,helber2019eurosat} ($5$ binary tasks) and \textit{\splitresisc}~\cite{cheng2017remote}, which divides $45$ classes into $9$ tasks.
    \item \textbf{Medical domain}: we adopt three settings, ranging from plant to human diseases. \textit{\splitcropdiseases}~\cite{hughes2015open} regards healthy/infected leaves with $7$ tasks/$5$ classes each. In \textit{\splitisic}~\cite{codella2018skin}, images of $6$ skin diseases are split into $3$ tasks. \textit{\splitchestx}~\cite{wang2017chestxray} consists of chest X-ray images ($2$ tasks/$3$ classes each).
\end{itemize}

\tit{Metrics.}~We report the \textbf{Final Average Accuracy} (FAA) -- computed after the end of the last task -- and the \textbf{Final Forgetting}~\cite{chaudhry2018riemannian} in suppl. materials. Each experiment is repeated three times with varying class orders~\cite{zhang2023slca}. We hence provide both the mean and the standard deviation of the FAA.

\smallskip
\tit{Implementation details.} We train our model with Adam~\cite{kinma2015adam} with a learning rate of $0.001$. The number of epochs is set according to the varying dataset sizes (see suppl.\ materials), while we adopt a batch size of $16$ for \splitimagenet and $128$ for the other datasets. For a fair comparison, we maintain the same number of epochs for all the competitors and search their optimal hyper-parameters (including learning rate and optimizer) to guarantee the best performance. These configurations are reported in suppl. materials. In line with most CL literature, we perform inference in an \textbf{instance-wise setup}, where each prompt is selected independently for each sample of the batch (more in suppl.\ materials). 

\smallskip
\tit{Competiting methods.}~We focus our evaluation on the state-of-the-art prompt tuning methods, including L2P~\cite{wang2022l2p}, DualPrompt~\cite{wang2022dualprompt}, AttriCLIP~\cite{wang2023attriclip}, PromptFusion~\cite{chen2023promptfusion}, and CODA-Prompt~\cite{smith2023coda}. Furthermore, our comparison includes methods that fine-tune all the network, including SLCA~\cite{zhang2023slca}, \dpp~\cite{buzzega2020dark} (\textit{rehearsal-based}), GDumb~\cite{prabhu2020gdumb} (\textit{rehearsal-based}), and LwF~\cite{li2017learning} (\textit{regularization-based}). We follow~\cite{wang2022dualprompt,smith2023coda,zhang2023slca} and adopt the same backbone based on \backbone as the main classification architecture for all the methods and settings (in our notation, $f(\cdot)$), initializing the network with a fully supervised pre-train on ImageNet-21K. Similarly, for all methods using CLIP, we use \clipback as in~\cite{wang2023attriclip} (in our notation, $E_{vis}(\cdot)$ and $E_{txt}(\cdot)$). We stress that our comparison comprehends methods such as PromptFusion and AttriCLIP that leverage the CLIP model in CL scenarios; we retain this ensures the fairness of our evaluation.

\smallskip
\tit{Training-free baselines and lower/upper bounds.}~We integrate our analysis with the zero-shot testing of CLIP~\cite{radford2021learning} (\textbf{Zero-shot CLIP}), which uses hand-crafted textual prompts (\textit{e.g.}, ``A photo of a $\operatorname{[CLS-NAME]}$'') to classify a testing image through~\cref{eq.1-stage-posterior}. Notably, the comparison between \methnam and the Zero-shot CLIP test provides an understanding of the algorithmic merits of our approach, and allows us to disentangle them from the strengths of CLIP. Futhermore, we consider a similar baseline built on top of the ImageNet pre-trained backbone. It freezes the network and simply memorize the latent features of each training example, employing a \textbf{k-NN} approach.
\input{tables/results_table}

For a thorough comparison, we include the results of \methnam when trained on all tasks in a stationary scenario -- \textbf{Joint (\textit{\methnam})} -- which serves as an upper bound. Moreover, we include \textbf{Joint~\textsuperscript{\textdagger} (\textit{\backbone})} and \textbf{Fine-tune~\textsuperscript{\textdagger} (\textit{\backbone})}, where a \backbone is fine-tuned on all tasks respectively jointly and incrementally with no countermeasure to forgetting.
\subsection{Comparison with the State of the Art}
\input{tables/results_table_ood}
As shown in Tab.~\ref{tab:main_results} and~\ref{tab:ood_results}, \methnam outperforms all the other approaches on average. The comparison is particularly significant when referring to other CLIP-based methods (\eg AttriCLIP and PromptFusion). For instance, in the two fine-grained scenarios -- \shortsplitcars and \shortsplitcub -- we get a $+11.16$ and a $+22.96$ margin over Zero-shot CLIP, respectively. We ascribe this result to the fact that fine-grained class differences (distinguishing a \qmarks{red headed woodpecker} from a \qmarks{red bellied woodpecker}) are not easily captured by the pre-trained CLIP. Therefore, a mechanism devoted to plasticity as our second-level prompting stage becomes fundamental. A similar conclusion could be drawn for the aerial and medical domains, where the average gap \wrt the Zero-shot CLIP is $+37.75$.

Remarkably, we get a $+5.96$ average boost over the runner-up method (SLCA), which is remarkable considering that \methnam has only a fraction of its learnable parameters (see suppl. materials). We also emphasize that \methnam closely matches the performance of its Joint upper bound, with an average margin of just $-3.59$. This result looks particularly promising: our solution yields an \textit{almost negligible drop in performance} \wrt a stationary scenario. Such an outcome holds true for both aerial and medical scenarios, which demand high adaptability.

Moreover, existing prompting approaches struggle in aerial and medical settings. They are surpassed by replay-based approaches, which can deeper fine-tune the backbone to account for domain shifts. Notably, \methnam stands out as the best compromise between stability and plasticity.

%% file: tables/results_table.tex
\begin{table*}[t]
    \caption{The Final Avg. Accuracy and std dev. for \textbf{natural} and \textbf{specialized} domains. \textsuperscript{\textdagger} denotes methods fine-tuning the whole model. \textsuperscript{*} highlights rehearsal approaches. We take results \textsuperscript{(1)} from~\cite{smith2023coda}, \textsuperscript{(2)} from~\cite{wang2023attriclip}, \textsuperscript{(3)} from~\cite{zhang2023slca}, \textsuperscript{(4)} from~\cite{chen2023promptfusion}. Due to the absence of a public codebase, we could not reproduce the results of PromptFusion on all datasets.}
    \centering
\rowcolors{2}{lightgray}{}
\begin{tabular}{lccccc}
\midrule
\textbf{Model}                                              & \textbf{\shortsplitimagenet}       & \textbf{\shortsplitcifar}                       & \textbf{\shortsplitcars}          & \textbf{\shortsplitcub} & \textbf{\textit{Avg.}}\\
\midrule
Joint (\textit{\methnam})                                                          & \result{90.03}{0.33}{-}       & \result{92.32}{0.28}{-}                    & \result{89.00}{0.33}{-}      & \result{85.64}{0.30}{-}  & \faa{89.25}\\
Joint~\textsuperscript{(3) \textdagger} (\textit{\backbone})     & \result{79.60}{0.87}{-}       & \result{93.22}{0.16}{-}                    & \result{80.31}{0.13}{-}      & \result{88.00}{0.34}{-} & \faa{85.28} \\
Fine-tune~\textsuperscript{\textdagger} (\textit{\backbone})             & \result{17.21}{4.72}{?}       & \result{17.47}{2.51}{?}                    & \result{9.28}{0.31}{?}       & \result{11.36}{1.43}{?}  & \faa{13.83}\\
k-NN (\textit{\backbone}) & \faa{18.93} & \faa{26.14} & \faa{11.57} & \faa{21.44} & \faa{19.52} \\
Zero-shot CLIP~\cite{radford2021learning}                   & \faa{85.36}       & \faa{73.27}                    & \faa{76.46}          & \faa{61.14}  & \faa{74.06}\\
\midrule
LwF~\textsuperscript{\textdagger}~\cite{li2017lwf}           & \result{19.09}{5.72}{?}       & \result{19.68}{0.90}{?}                    & \result{23.24}{1.88}{?}      & \result{16.73}{4.16}{?}     & \faa{19.69}\\
GDumb~\textsuperscript{* \textdagger}~\cite{prabhu2020gdumb}  & \result{44.28}{0.51}{?}              & \result{57.92}{1.67}{?}                    & \result{28.74}{0.47}{?}      & \result{61.34}{0.46}{?}  & \faa{48.07} \\
\dpp~\textsuperscript{* \textdagger}~\cite{buzzega2020dark}   & \result{56.66}{0.97}{?}              & \result{79.77}{1.14}{?}                    & \result{53.66}{1.51}{?}      & \result{74.62}{0.73}{?}  & \faa{66.18}\\
L2P~\textsuperscript{(3)}~\cite{wang2022l2p}                                      & \result{66.49}{0.40}{72.83}   & \result{82.76}{1.17}{88.48}                & \result{38.18}{2.33}{51.79}  & \result{62.21}{1.92}{73.83}  & \faa{62.41}\\
DualPrompt~\textsuperscript{(3)}~\cite{wang2022dualprompt}                        & \result{68.50}{0.52}{72.59}   & \result{85.56}{0.33}{90.33}                & \result{40.14}{2.36}{56.74}  & \result{66.00}{0.57}{77.92}  & \faa{65.05}\\
CODA-Prompt~\cite{smith2023coda}                            & \resultC{75.45}{0.56}{1.64}{1}    & \resultC{86.25}{0.74}{1.67}{1}                 & \result{31.99}{3.39}{}       & \result{67.30}{3.19}{}  & \faa{65.25}\\
AttriCLIP~\cite{wang2023attriclip}                          & \result{86.25}{0.75}{}       & \resultsC{81.40}{0.12}{}{2}                      & \result{70.98}{0.41}{}                & \result{50.07}{1.37}{}  & \faa{72.18}\\
PromptFusion~\textsuperscript{(4) *}~\cite{chen2023promptfusion}                          & \result{80.70}{-}{}        & \result{87.40}{-}{}                      & $-$                & $-$  & $-$ \\
SLCA~\textsuperscript{(3) \textdagger}~\cite{zhang2023slca}      & \result{77.00}{0.33}{81.17}   & \resultb{91.53}{0.28}{94.09}               & \result{67.73}{0.85}{76.93}  & \resultb{84.71}{0.40}{90.94}  & \faa{80.24}\\
\midrule
\textbf{\methnam}    & \resultb{89.83}{0.04}{3.62}   & \result{90.12}{0.32}{0.51}     & \resultb{87.62}{0.20}{6.11}  & \result{84.10}{0.28}{6.88} & \faab{87.92} \\
\midrule
\end{tabular}
\vspace{-1em}
    \label{tab:main_results}
\end{table*}

%% file: tables/results_table_ood.tex
\begin{table*}[t]
    \caption{The Final Avg. Accuracy for the \textbf{aerial} and \textbf{medical} domains. The std dev. is reported in the suppl. materials for the sake of presentation.}
    \centering
    {
\centering
\rowcolors{2}{lightgray}{}
\begin{tabular}{lccccccc}
\midrule
\textbf{\small{Model}} & \textbf{\shortspliteurosat} & \textbf{\shortsplitresisc} & \textbf{\shortsplitcropdiseases} & \textbf{\shortsplitisic} & \textbf{\shortsplitchestx} & \textbf{\textit{Avg.}} \\
\midrule
Joint (\textit{\methnam}) & \resultNV{97.21}{0.13}{-} & \resultNV{96.41}{0.24}{-} & \resultNV{99.19}{0.10}{-} & \resultNV{78.25}{0.30}{-} & \resultNV{45.36}{0.43}{-}  & \faa{83.28}\\
Joint~\textsuperscript{\textdagger} (\textit{\backbone}) & \resultNV{98.19}{0.12}{} & \resultNV{96.88}{0.10}{} & \resultNV{99.68}{0.08}{} & \resultNV{88.31}{1.76}{} & \resultNV{48.92}{1.40}{} & \faa{86.40} \\
Fine-tune~\textsuperscript{\textdagger} (\textit{\backbone}) & \resultNV{19.91}{0.13}{} & \resultNV{14.96}{2.02}{} & \resultNV{13.24}{0.39}{} & \resultNV{30.30}{2.22}{} & \resultNV{30.92}{0.64}{} & \faa{21.87} \\
k-NN (\textit{\backbone}) & \faa{28.18} & \faa{24.86} & \faa{30.74} & \faa{19.09} & \faa{11.51}  & \faa{22.88} \\
Zero shot CLIP & \faa{54.50} & \faa{63.15} & \faa{27.58} & \faa{29.14} & \faa{26.27} & \faa{40.13} \\
\midrule
LwF~\textsuperscript{\textdagger} & \resultNV{25.13}{2.78}{} & \resultNV{15.37}{0.85}{} & \resultNV{22.31}{2.76}{} & \resultNV{33.06}{1.98}{} & \resultNV{32.82}{0.85}{} & \faa{25.74} \\
GDumb~\textsuperscript{* \textdagger} & \resultNV{90.99}{1.49}{} & \resultNV{60.07}{0.54}{} & \resultNV{83.61}{1.35}{} & \resultNV{61.64}{3.64}{} & \resultNV{32.33}{2.26}{} & \faa{65.73} \\
\dpp~\textsuperscript{* \textdagger} & \resultNV{93.08}{1.62}{} & \resultNV{51.84}{2.89}{} & \resultNV{92.53}{1.06}{} & \resultNV{65.68}{2.16}{} & \resultNV{35.52}{1.22}{} & \faa{67.72} \\
L2P & \resultNV{46.34}{7.86}{} & \resultNV{63.27}{3.71}{} & \resultNV{74.68}{0.25}{} & \resultNV{47.13}{3.84}{} & \resultNV{32.46}{1.52}{}  & \faa{52.78} \\
DualPrompt & \resultNV{71.39}{4.94}{} & \resultNV{76.21}{3.92}{} & \resultNV{81.41}{2.68}{} & \resultNV{49.99}{1.07}{} & \resultNV{35.70}{0.10}{} & \faa{62.94} \\
CODA-Prompt & \resultNV{63.12}{6.30}{} & \resultNV{70.46}{5.15}{} & \resultNV{77.09}{2.91}{} & \resultNV{44.87}{3.50}{} & \resultNV{38.62}{3.90}{} & \faa{58.83} \\
AttriCLIP & \resultNV{57.51}{6.27}{} & \resultNV{66.64}{2.48}{} & \resultNV{33.21}{7.14}{} & \resultNV{26.77}{11.09}{} & \resultNV{28.94}{1.27}{} & \faa{42.61} \\
SLCA~\textsuperscript{\textdagger} & \resultNV{88.69}{0.48}{} & \resultNV{85.70}{0.35}{} & \resultNV{93.80}{0.60}{} & \resultNV{59.19}{3.83}{} & \resultNV{39.07}{1.80}{} & \faa{73.29} \\
\midrule
\textbf{\methnam} & \resultbNV{93.70}{0.15}{5.12} & \resultbNV{92.28}{0.54}{5.12} & \resultbNV{94.92}{0.60}{3.2} & \resultbNV{66.67}{1.45}{26.36} & \resultbNV{41.85}{2.63}{30.75} & \faab{77.88} \\
\midrule
\end{tabular}
}
    \label{tab:ood_results}
\vspace{-0.8em}
\end{table*}

%% file: sec/5_ablations.tex
\subsection{Ablation studies}
\label{sec:ablations}
\begin{table*}[t]
    \centering
    \caption{Ablative studies on \methnam (Final Avg. Acc. $\pm$ std dev).}
    \input{tables/results_concat_vs_sum}
    \label{tab:abl_results}
\end{table*}
\cref{tab:abl_results} dissects the impact of each part of \methnam. For space constraints, we report one dataset for each domain (see suppl. materials for the others).

\tit{Two-level prompting.}~Through the first two comparisons, we assess the role of the two-level strategy. Specifically, \textit{\qmarks{Classify with first-level keys $\pmb{w}_c$}} indicates an ablative approach that lacks the second stage. To classify, similarly to CoOp, it uses the learned keys $\pmb{w}_c$ as class-prototypes (the posteriors are computed as in~\cref{eq.1-stage-posterior}). Tab.~\ref{tab:abl_results} shows that this strategy achieves competitive performance, highlighting the stability of the learned keys. However, if more plasticity is required (as for \shortspliteurosat and \shortsplitisic), the gap w.r.t. \methnam increases. Similarly, \methnam shows improvement over the row labeled \textit{\qmarks{w/o first-level prompts}}, which is a variation of \methnam that avoids learning fist-level prompts and replaces them with the static hand-crafted CLIP textual templates~\cite{radford2021learning}. We use the resulting textual embeddings as class-prototype keys, learning only second-level prompts $Q_c$.

\tit{Semantic residuals.}~Afterwards, we replace our semantic residual (\cref{eq.residual-in-ViT-MSA}) with \textit{Prefix Tuning}: following~\cite{wang2022dualprompt}, for each ViT layer we train $5$ key and $5$ value prompt tokens, concatenated to the keys and the values of the respective MSA. Still, the results in \cref{tab:abl_results} highlight the advantage of our additive mechanism.

\tit{Generative Replay.}~A significant drop occurs if the generative replay stage is removed, especially when tasks deviate from the pre-train. Moreover, row \textit{w.\ Unimodal Generative Replay} shows the results obtained when using a single Gaussian instead of a MoG. Despite it is generally better than \textit{w/o Generative Replay}, the results are largely inferior to the MoG used for the full model.

\tit{Confidence modulation of the semantic residuals.}~\cref{eq.semantic-residual} injects the \textit{confidence} of the CLIP encoders into the residuals. The last row shows that \textit{Confidence Modulation} provides a fair improvement across the different datasets.

%% file: tables/results_concat_vs_sum.tex
\centering
\begin{tabular}{lcccc}
\midrule
\textbf{Model} & \textbf{\shortsplitimagenet} & \textbf{\shortsplitcars} & \textbf{\shortspliteurosat} & \textbf{\shortsplitisic} \\
\midrule
\textbf{\methnam} & \resultb{89.83}{0.04}{4.14} & \resultb{87.62}{0.20}{6.78} & \resultb{93.70}{0.09}{} & \resultb{66.67}{1.45}{26.36}  \\
\midrule
\multicolumn{5}{c}{\textbf{Ablations on two-level prompting}} \\
\midrule
\rowcolor{lightgray}
\textit{Classify with first-level keys $\pmb{w}_c$} & \result{88.16}{0.27}{3.88} & \result{87.57}{0.12}{4.78} & \result{90.12}{0.54}{5.32} & \result{58.87}{1.25}{21.64} \\
\textit{w/o first-level prompts} & \result{88.97}{0.52}{} & \result{79.88}{0.46}{13.50} & \result{86.25}{5.32}{} & \result{58.68}{5.57}{} \\
\midrule
\multicolumn{5}{c}{\textbf{Other secondary ablations}} \\
\midrule
\rowcolor{lightgray}
\textit{Prefix Tuning (no residuals)} & \result{71.34}{0.34}{6.52} & \result{60.03}{4.14}{} & \result{90.92}{0.44}{3.90} & \result{62.46}{0.88}{20.03} \\
\textit{w/o Generative Replay} & \result{88.55}{0.06}{4.42} & \result{87.17}{0.07}{} & \result{83.78}{1.82}{} & \result{50.76}{3.36}{} \\
\rowcolor{lightgray}
\textit{w.\ Unimodal Generative Replay} & \result{89.62}{0.12}{} & \result{87.28}{0.12}{} & \result{92.58}{0.40}{} & \result{60.79}{0.68}{} \\
\textit{w/o Confidence Modulation} & \result{88.73}{0.16}{} & \result{87.29}{0.13}{} & \result{93.68}{0.38}{} & \result{63.53}{0.75}{} \\
\midrule
\end{tabular}
\vspace{-0.8em}

%% file: sec/6_conclusions.tex
\section{Conclusion}
We present \methnam, a prompting method for Continual Learning based on three novelties. First, we strengthen the stability of prompt selection using a foundation model and two levels of prompt tuning. Second, we replace standard prompt concatenation with additive residuals, which transfer semantics into MLP layers. Finally, we use a simple generative replay based on a multi-modal representation of the feature distributions. Each part of \methnam brings a significant  contribution, leading it to outperform the state of the art.

%% file: suppl_arxiv.tex
\clearpage
\setcounter{page}{1}
\setcounter{figure}{0}
\setcounter{equation}{0}
\setcounter{table}{0}
\pagenumbering{roman}

\renewcommand{\thetable}{\Alph{table}}
\renewcommand{\theequation}{\Alph{equation}}
\renewcommand{\thefigure}{\Alph{figure}}

\appendix

\section{Prompt-selection stability}
\label{suppl:prompt_selection_stability}
We hereby extend the results outlined in \cref{sec:discussion} and present outcomes for additional datasets/domains. \Cref{fig:cf_suppl} provides a visual snapshot at the conclusion of the final task. It is evident that \methnam exhibits the highest precision in selecting the appropriate prompts. \Cref{fig:f1_suppl} supplements this analysis with a detailed focus on the initial task. At first glance, CODA-Prompt~\cite{smith2023coda} appears to match the retrieval performance of our approach in \spliteurosat. However, we ascribe it to a bias of CODA-Prompt towards the prompts learned during the first task, as also observed in \cref{fig:cf_suppl}. These additional results reaffirm our belief that \methnam strikes the optimal balance between stability and flexibility throughout the training sequence.

\input{figures/retrieval_acc_suppl}

\section{Additional methodological details}
\label{suppl:method_details}
In this section, we provide the details concerning ${\cal L}_{\operatorname{GR}}^{Q}$, introduced in
\cref{subsec:mogs}, and which closely follows the computation of ${\cal L}_{\operatorname{GR}}^{P}$. 
Specifically, similarly to \cref{eq.cross-entropy-reharsal},
${\cal L}_{\operatorname{GR}}^{Q}$ is defined as:
\begin{equation}
\label{eq.cross-entropy-reharsal-Q}
{\cal L}_{\operatorname{GR}}^Q = - \frac{1}{n Nt}
 \sum_{c=1}^{Nt} \sum_{i=1}^n \log p(y_i = c|\tilde{\pmb{e}}_{L,i}^{\operatorname{CLS}}),
\end{equation}

\noindent where $p(y_i = c|\tilde{\pmb{e}}_{L,i}^{\operatorname{CLS}})$ is the score relating each synthetic representation $\tilde{\pmb{e}}_{L,i}^{\operatorname{CLS}} \sim \mathcal{H}_{c}$ with the corresponding ground-truth class $c$. We compute it as:

\begin{equation}
\label{eq.2-stage-posterior-reharsal}
 p(y_i = c|\tilde{\pmb{e}}_{L,i}^{\operatorname{CLS}})= g_{\theta_{t'} \vert_{t' \in \{ 1, ..., t \} }} (\tilde{\pmb{e}}_{L,i}^{\operatorname{CLS}}).
\end{equation}

In \cref{eq.2-stage-posterior-reharsal}, $g_{\theta_{t'} \vert_{t' \in \{ 1, ..., t \} }} (\cdot)$ indicates that we use {\em all} the classification heads of $g(\cdot)$ corresponding to the $t$ tasks so far observed.

\input{figures/first_task_acc_suppl}

\section{On the computational demands of prompt-based approaches}
\label{suppl:computational_demand}
We use a serial two-stage training approach. First, we learn prompts for the CLIP text encoder, freeze them, and then learn prompts for the ImageNet-pre-trained ViT. Since each phase is independent, the overall cost stems from each respecting phase.

In the first stage, similar to~\cite{zhou2022learning}, a batch of images requires one forward pass through the frozen CLIP image encoder and another through the text encoder to compute class-level textual embeddings. Due to the very low number of trainable parameters, this stage converges in a few epochs and requires no gradient computation for the image encoder.

In the second stage, we freeze the CLIP text encoder prompts and focus on the ViT prompts. The class prototypes from the text encoder can be cached, eliminating the need for additional forward passes through the text encoder. Again, each batch involves two forward passes: one through the CLIP image encoder and another through the prompted ViT. This complexity of \methnam thus matches L2P, Dual-Prompt, and CODA-Prompt, which also use two forward passes but with the same backbone. Our approach uses two distinct backbones, requiring additional GPU memory for the CLIP visual encoder, which we find negligible compared to the performance gain.

\section{Setting and implementation details}
\label{suppl:dataset_and_setting}
\tit{Experimental setting.} We here provide any missing information about datasets and experimental settings:
\begin{itemize}
    \item \textbf{\splitimagenet}: $10$ tasks of $20$ classes each; $50$ training epochs.
    \item \textbf{\splitcifar}: $10$ tasks of $10$ classes each; $20$ training epochs.
    \item \textbf{\splitcars}: $9$ tasks of $20$ classes each, plus the 10th task comprising the remaining 16 classes; $50$ training epochs.
    \item \textbf{\splitcub}: $10$ tasks of $20$ classes each; $50$ training epochs.
    \item \textbf{\spliteurosat}: $5$ tasks of $2$ classes each; $5$ training epochs.
    \item \textbf{\splitresisc}: $9$ tasks of $5$ classes each; $30$ training epochs.
    \item \textbf{\splitcropdiseases}: $7$ tasks of $5$ classes each; $5$ training epochs. From the original dataset~\cite{hughes2015open} of $38$ classes, we removed the classes with the lowest number of examples (\qmarks{\textit{Potato healthy}}, \qmarks{\textit{Apple Cedar}} and \qmarks{\textit{Peach healthy}}).
    \item \textbf{\splitisic}: $3$ tasks of $2$ classes each; $30$ training epochs. From the original dataset~\cite{codella2018skin} we removed the most frequent class \qmarks{\textit{Melanocytic nevus}}.
    \item \textbf{\splitchestx}: $2$ tasks of $3$ classes each. $30$ training epochs. From the original dataset~\cite{wang2017chestxray}, we took the images without overlapping diseases belonging to the classes \qmarks{\textit{Cardiomegaly}}, \qmarks{\textit{Consolidation}}, \qmarks{\textit{Edema}}, \qmarks{\textit{Fibrosis}}, \qmarks{\textit{Pleural Thickening}} and \qmarks{\textit{Pneumothorax}}.
\end{itemize}
Following~\cite{zhang2023slca}, each experiment is repeated $3$ times fixing the seeds $1993$, $1996$, and $1997$.

\tit{On instance/batch-wise inference.} To maintain consistency with the majority of CL studies, we conduct inference independently for each sample within a batch. This approach, termed \textit{instance-wise} inference, is by far the most predominant in the literature~\cite{li2017learning,rebuffi2017icarl,aljundi2019gradient,zhang2023slca,smith2023coda}. 

Conversely, L2P and DualPrompt originally presented results using a \textit{batch-wise} setup, where a single prompt is selected for all samples in a batch through majority voting. We reckon that this setup offers an unfair advantage, as it leverages the fact that samples are not shuffled during inference, hence the ground-truth labels of the examples within a mini-batch are typically the same. To ensure a fair comparison with other methods, we thus report L2P and DualPrompt results using the instance-wise setup, thereby eliminating any potential advantage these approaches might have had over other techniques.

\section{Final Forgetting Metric}
\label{suppl:forgetting}
\cref{tab:forg_results,tab:ood_forgetting} report the Final Forgetting metric~\cite{chaudhry2018riemannian} for our experiments. \cref{tab:forg_results} lacks L2P, DualPrompt, and SLCA, since~\cite{zhang2023slca} does not report forgetting values for their experiments. The same applies to PromptFusion and to the experiments of AttriCLIP on \splitcifar (taken from~\cite{wang2023attriclip}).

Moreover, we report in \cref{tab:ood_stddev} the standard deviation of experiments in \cref{tab:ood_results} of the main paper.

\input{tables/forgetting}
\input{tables/forgetting_ood}
\input{tables/stddev_ood}
\section{Hyperparameters}
\label{suppl:hyperparams}
The hyperparameters of \methnam employed for each experiment are reported in \cref{tab:hyperparams,tab:hyperparams_ood}.

\input{sec/suppl/hyperparams}
\input{sec/suppl/hyperparams_ood}
\section{Number of Trainable Parameters} 
\label{suppl:number_of_parameters}
The number of trainable parameters varies a lot among the compared methods. As mentioned in \cref{sec:intro,sec:related}, the two main adaptation strategies are \textbf{fine-tuning} the whole model on the training data of the target dataset(s) or \textbf{parameter-efficient learning}, which adapts the model with only a few parameters (e.g., the prompts). In \cref{tab:tab_params_abl} we use \splitcifar as reference scenario and report the number of trainable parameters, \ie, those parameters that are optimized during the incremental learning. For instance, SLCA fine-tunes the whole model, while the trainable parameters of STAR-Prompt are composed of: $\pmb{p}_c, Q_c, A_c$ ($c \in {\cal Y}_1, ..., {\cal Y}_T$) and $\theta_t$ ($t \in \{1, ..., T \}$).

\input{tables/params}

\section{Additional ablations}
\label{suppl:ablations_other_datasets}
We herein report the ablative studies of \cref{sec:ablations} over the remaining datasets in \cref{tab:abl_suppl}. These results confirm the conclusion already outlined in the main paper (see \cref{sec:ablations}).

\input{tables/result_abl_suppl}
\section{Additional experimental analysis}
\label{suppl:more_ablations}

\tit{Number of Gaussians.} \cref{tab:tab_m_abl} shows the Final Average Accuracy of STAR-Prompt on Split Imagenet-R using different numbers of Gaussians ($M$) for each $\mathcal{H}_{c}$. The results indicate a substantial plateau of the performance when $M \geq 5$, which we hence set as default value in all our experiments. 

\tit{Prefix tuning \textit{vs.} semantic residuals.} In \cref{tab:results_c_vs_s} we extend the comparison between prefix tuning and semantic residuals reported in \cref{tab:abl_results} of the main paper. Specifically, differently from \textit{Prefix Tuning} in \cref{tab:abl_results}, in which we used 5 tokens for each key and value, in \cref{tab:results_c_vs_s} we use \textit{one} token only for each key and value. This way, the total number of parameters per class is comparable with $Q_c$.
\cref{tab:results_c_vs_s} shows that the Final Average Accuracy scores obtained using prefix tuning are largely inferior to using semantic residuals, confirming the results reported in \cref{tab:abl_results}.
This shows that the gap between the two conditioning methods is not due to the number of prompt parameters. The results of prefix tuning in \ref{tab:results_c_vs_s} are drastically inferior to those obtained when using the recipe suggested in~\cite{wang2022dualprompt} (5 tokens for each key and value) and adopted in \cref{tab:abl_results}.

\input{tables/n_gaussians_table}

\input{tables/abl_concat_vs_sum}

%% file: figures/retrieval_acc_suppl.tex
\begin{figure}[h]
\begin{minipage}[t]{\linewidth}
    \centering
    \renewcommand{\arraystretch}{-4.5}
    \setlength{\tabcolsep}{2pt}
    \begin{tabular}{cc}
    \includegraphics[width=0.49\textwidth]{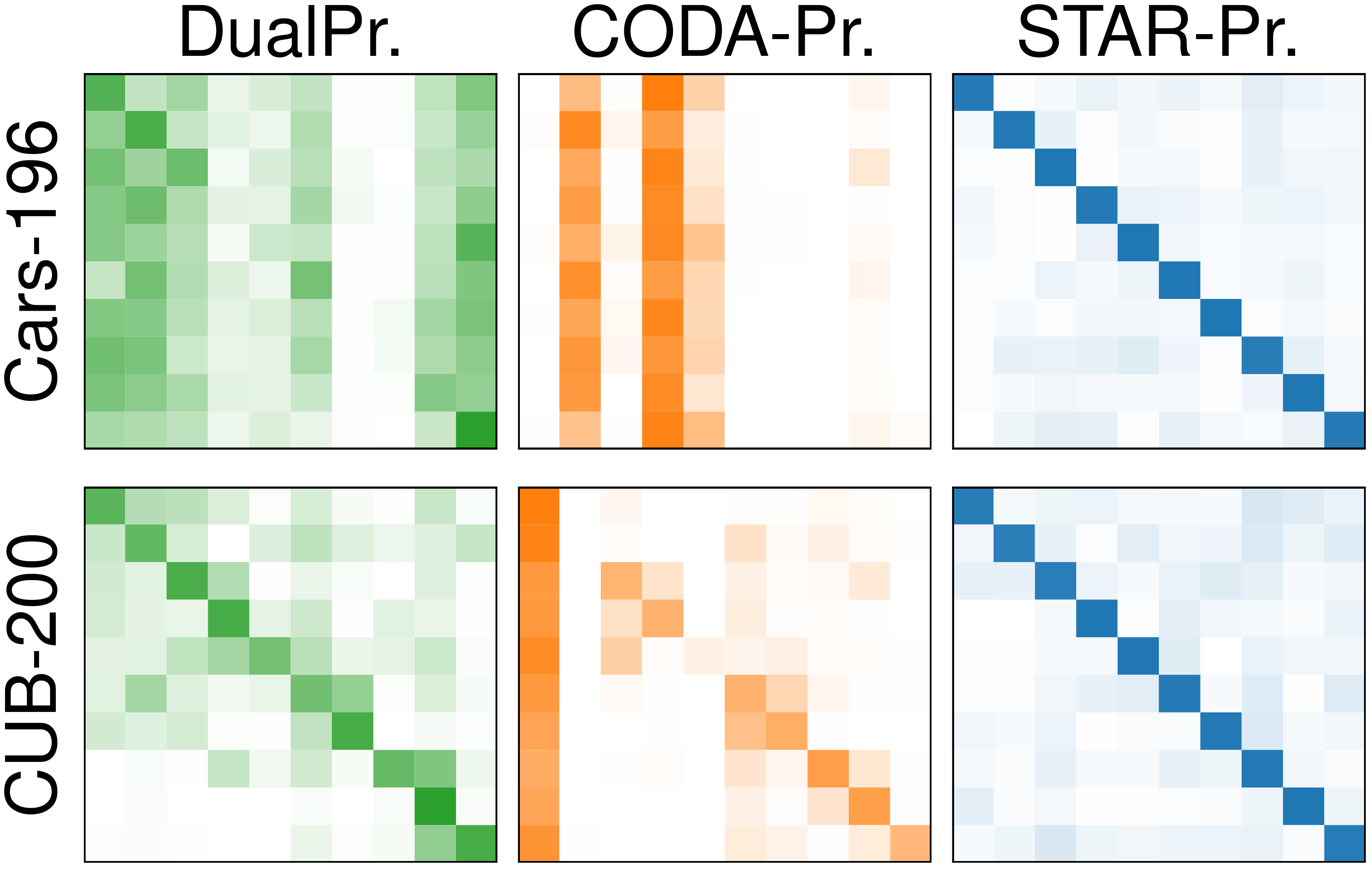} &
    \includegraphics[width=0.49\textwidth]{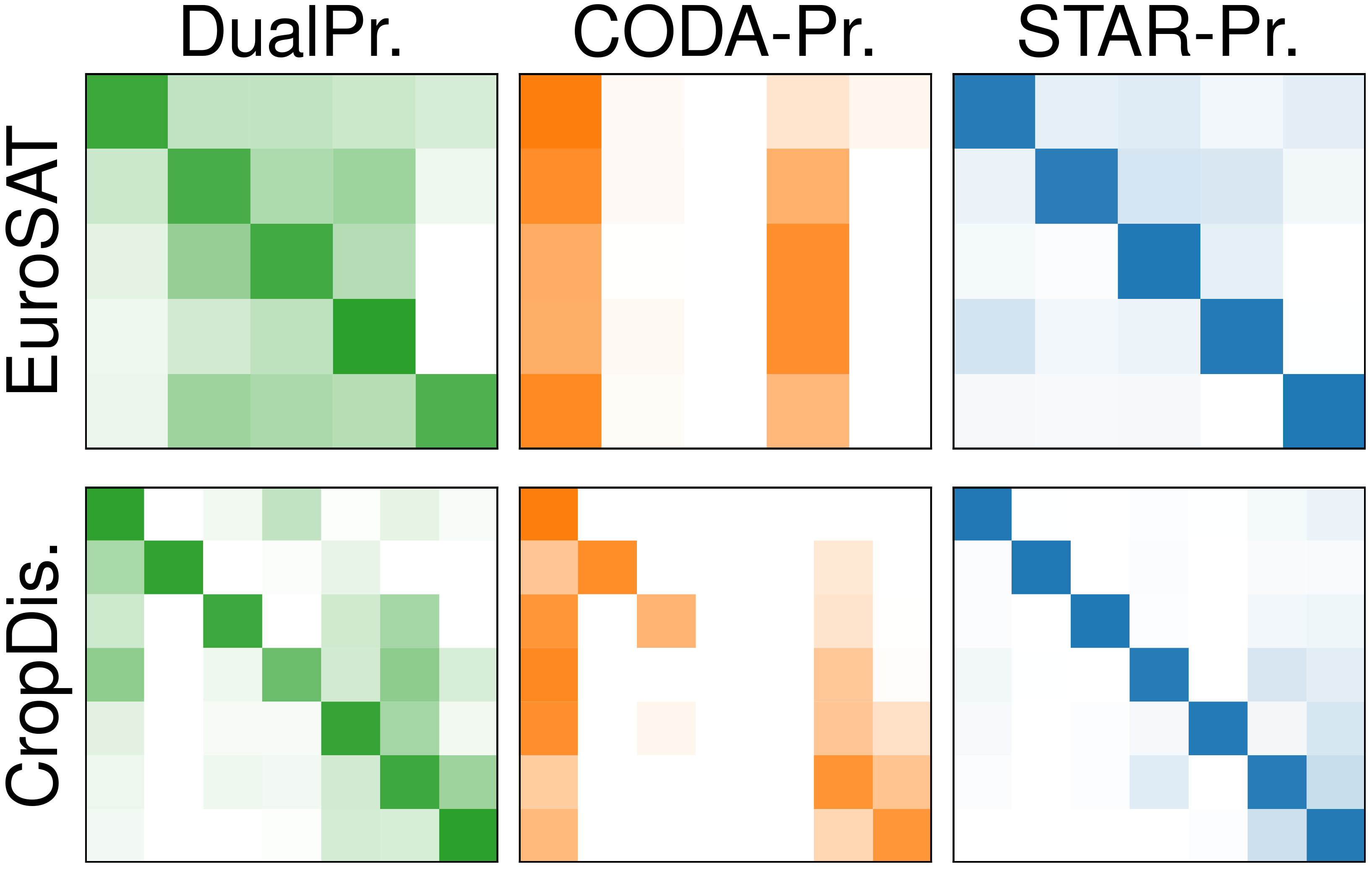} \\
    \end{tabular}
\end{minipage} 
    \caption{Analysis of prompt selection stability for \splitcars, \spliteurosat, \splitcub, and \splitcropdiseases. We assess various models at the end of the last task and report results as confusion matrices. The $y$ axis indicates the task of the query sample, while the $x$ axis shows the task of the corresponding selected key.}
    \label{fig:cf_suppl}
\end{figure}

%% file: figures/first_task_acc_suppl.tex
\begin{figure*}[t]
    \centering
    \begin{tabular}{cc}
    \multicolumn{2}{c}{
    \includegraphics[width=0.50\textwidth]{imgs/f1legend.pdf}
    } \\
    \includegraphics[width=0.48\textwidth]{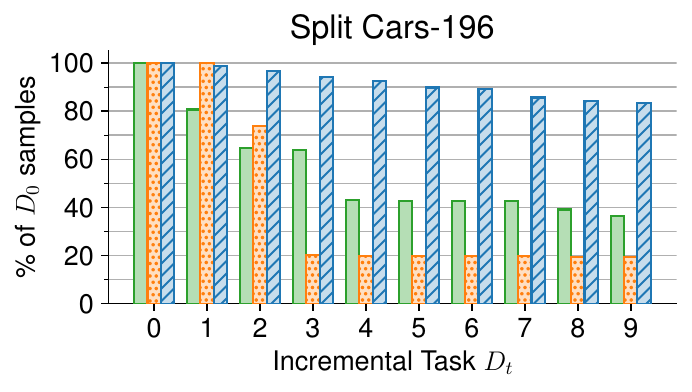} &
    \includegraphics[width=0.48\textwidth]{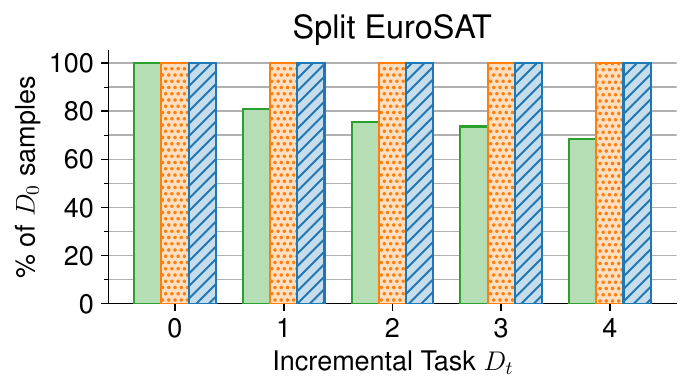} \\
    \end{tabular}
        
    \caption{Prompt retrieval on the first task of different datasets.}
    \label{fig:f1_suppl}
\end{figure*}

%% file: tables/forgetting.tex
\begin{table*}[t]
    \caption{The Final Forgetting (\textit{the lower the better}) - \textbf{part 1}}

\centering
\rowcolors{2}{lightgray}{}
\begin{tabular}{lccccc}
\toprule
\textbf{Model}                                              & \textbf{\shortsplitimagenet}       & \textbf{\shortsplitcifar}                       & \textbf{\shortsplitcars}          & \textbf{\shortsplitcub} & \textbf{\textit{Avg.}} \\
\midrule
Fine-tune~\textsuperscript{\textdagger} (\textit{\backbone})             & \resultNV{83.62}{2.80}{?}       & \resultNV{89.95}{2.63}{?}                    & \resultNV{85.55}{0.99}{?}       & \resultNV{90.69}{3.34}{?} & \faa{87.45} \\
\midrule
LwF~\textsuperscript{\textdagger}           & \resultNV{79.68}{4.64}{?}       & \resultNV{87.23}{0.91}{?}                    & \resultNV{76.59}{1.65}{?}      & \resultNV{83.81}{6.71}{?} & \faa{81.83}     \\
\dpp~\textsuperscript{* \textdagger}   & \resultNV{38.88}{1.07}{?}              & \resultNV{19.28}{1.18}{?}                    & \resultNV{24.42}{0.22}{?}      & \resultNV{19.16}{1.08}{?} & \faa{25.44} \\
CODA-Prompt                            & \resultNV{4.06}{0.37}{}    & \resultNV{5.56}{0.72}{}                 & \resultNV{7.36}{2.74}{}       & \resultbNV{5.65}{1.48}{} & \faa{5.66} \\
AttriCLIP                         & \resultNV{6.9}{}{}       & $-$                      & \resultNV{18.8}{}{}                & \resultNV{33.35}{}{} & \faa{19.68} \\
\midrule
\textbf{\methnam}  & \resultbNV{3.92}{}{}   & \resultbNV{4.71}{}{}     & \resultbNV{6.14}{}{}  & \resultNV{6.94}{}{} & \faab{5.43} \\
\bottomrule
\end{tabular}

\label{tab:forg_results}
\end{table*}

%% file: tables/forgetting_ood.tex
\begin{table*}[t]
    \caption{The Final Forgetting (\textit{the lower the better}) - \textbf{part 2}}
    \centering
    {
\centering
\rowcolors{2}{lightgray}{}
\begin{tabular}{lccccccc}
\midrule
\textbf{\small{Model}} & \textbf{\shortspliteurosat} & \textbf{\shortsplitresisc} & \textbf{\shortsplitcropdiseases} & \textbf{\shortsplitisic} & \textbf{\shortsplitchestx} & \textbf{\textit{Avg.}} \\
\midrule
Fine-tune~\textsuperscript{\textdagger} (\textit{\backbone}) & \resultNV{98.11}{0.13}{} & \resultNV{95.23}{2.02}{} & \resultNV{93.07}{0.39}{} & \resultNV{94.31}{2.22}{} & \resultNV{66.76}{0.64}{} & \faa{78.23} \\
\midrule
LwF~\textsuperscript{\textdagger} & \resultNV{92.88}{2.78}{} & \resultNV{94.63}{0.85}{} & \resultNV{90.57}{2.76}{} & \resultNV{95.06}{1.98}{} & \resultNV{69.01}{0.85}{} & \faa{77.98} \\
\dpp~\textsuperscript{* \textdagger} & \resultNV{8.02}{1.62}{} & \resultNV{53.52}{2.89}{} & \resultNV{8.57}{1.06}{} & \resultNV{45.61}{2.16}{} & \resultNV{61.61}{1.22}{} & \faa{40.84} \\
L2P & \resultNV{43.87}{7.86}{} & \resultNV{25.41}{3.71}{} & \resultNV{18.85}{0.25}{} & \resultNV{37.79}{3.84}{} & \resultNV{42.60}{1.52}{}  & \faa{36.88} \\
DualPrompt & \resultNV{12.88}{4.94}{} & \resultNV{14.46}{3.92}{} & \resultNV{10.98}{2.68}{} & \resultbNV{25.01}{1.07}{} & \resultNV{31.35}{0.10}{} & \faa{26.27} \\
CODA-Prompt & \resultNV{16.75}{6.30}{} & \resultNV{15.05}{5.15}{} & \resultNV{10.39}{2.91}{} & \resultNV{22.97}{3.50}{} & \resultbNV{10.49}{3.90}{} & \faa{22.41} \\
AttriCLIP & \resultNV{39.13}{}{} & \resultNV{32.16}{}{} & \resultNV{62.56}{}{} & \resultNV{74.72}{}{} & \resultNV{48.57}{}{} & \faa{51.43} \\
SLCA~\textsuperscript{\textdagger} & \resultNV{7.74}{0.48}{} & \resultNV{10.58}{0.35}{} & \resultNV{4.90}{0.60}{} & \resultNV{35.25}{3.83}{} & \resultNV{32.05}{1.80}{} & \faa{27.30} \\
\midrule
\textbf{\methnam} & \resultbNV{4.31}{0.09}{} & \resultbNV{5.25}{0.17}{} & \resultbNV{3.26}{1.59}{} & \resultNV{26.36}{3.08}{} & \resultNV{30.75}{1.01}{} & \faab{14.61} \\
\midrule
\end{tabular}
}
    \label{tab:ood_forgetting}
\end{table*}

%% file: tables/stddev_ood.tex
\begin{table*}[t]
    \caption{The std dev. for experiments in \cref{tab:ood_results}.}
    \centering
    {
\centering
\rowcolors{2}{lightgray}{}
\begin{tabular}{lcccccc}
\midrule
\textbf{\small{Model}} & \textbf{\shortspliteurosat} & \textbf{\shortsplitresisc} & \textbf{\shortsplitcropdiseases} & \textbf{\shortsplitisic} & \textbf{\shortsplitchestx} \\
\midrule
Joint (\textit{\methnam}) & \juststd{}{0.13}{-} & \juststd{}{0.24}{-} & \juststd{}{0.10}{-} & \juststd{}{0.30}{-} & \juststd{}{0.43}{-} \\
Joint~\textsuperscript{\textdagger} (\textit{\backbone}) & \juststd{98.19}{0.12}{} & \juststd{96.88}{0.10}{} & \juststd{99.68}{0.08}{} & \juststd{88.31}{1.76}{} & \juststd{48.92}{1.40}{} \\
Fine-tune~\textsuperscript{\textdagger} (\textit{\backbone}) & \juststd{19.91}{0.13}{} & \juststd{14.96}{2.02}{} & \juststd{13.24}{0.39}{} & \juststd{30.30}{2.22}{} & \juststd{30.92}{0.64}{} \\
\midrule
LwF~\textsuperscript{\textdagger} & \juststd{25.13}{2.78}{} & \juststd{15.37}{0.85}{} & \juststd{22.31}{2.76}{} & \juststd{33.06}{1.98}{} & \juststd{32.82}{0.85}{} \\
GDumb~\textsuperscript{* \textdagger} & \juststd{90.99}{1.49}{} & \juststd{60.07}{0.54}{} & \juststd{83.61}{1.35}{} & \juststd{61.64}{3.64}{} & \juststd{32.33}{2.26}{} \\
\dpp~\textsuperscript{* \textdagger} & \juststd{93.08}{1.62}{} & \juststd{51.84}{2.89}{} & \juststd{92.53}{1.06}{} & \juststd{65.68}{2.16}{} & \juststd{35.52}{1.22}{} \\
L2P & \juststd{46.34}{7.86}{} & \juststd{63.27}{3.71}{} & \juststd{74.68}{0.25}{} & \juststd{47.13}{3.84}{} & \juststd{32.46}{1.52}{} \\
DualPrompt & \juststd{71.39}{4.94}{} & \juststd{76.21}{3.92}{} & \juststd{81.41}{2.68}{} & \juststd{49.99}{1.07}{} & \juststd{35.70}{0.10}{} \\
CODA-Prompt & \juststd{63.12}{6.30}{} & \juststd{70.46}{5.15}{} & \juststd{77.09}{2.91}{} & \juststd{44.87}{3.50}{} & \juststd{38.62}{3.90}{} \\
AttriCLIP & \juststd{66.00}{6.15}{} & \juststd{66.37}{4.31}{} & \juststd{32.75}{3.03}{} & \juststd{28.32}{10.38}{} & \juststd{24.75}{1.81}{} \\
SLCA~\textsuperscript{\textdagger} & \juststd{88.69}{0.48}{} & \juststd{85.70}{0.35}{} & \juststd{93.80}{0.60}{} & \juststd{59.19}{3.83}{} & \juststd{39.07}{1.80}{} \\
\midrule

\textbf{\methnam} & \juststd{93.70}{0.15}{5.12} & \juststd{92.28}{0.54}{5.12} & \juststd{94.92}{0.60}{3.2} & \juststd{65.38}{0.62}{33.02} & \juststd{41.85}{2.63}{30.75}\\

\midrule
\end{tabular}
}
    \label{tab:ood_stddev}
\end{table*}

%% file: sec/suppl/hyperparams.tex
\begin{table*}[t]
    \begin{center}
\rowcolors{2}{lightgray}{}
\begin{tabular}{lccccc}
\toprule
\textbf{} & \textbf{Refs. to Algorithm} & \textbf{\shortsplitimagenet} & \textbf{\shortsplitcifar} & \textbf{\shortsplitcars} & \textbf{\shortsplitcub} \\
\midrule
$E_1$ & \textit{First stage} -- Line 1 & $10$ & $10$ & $10$ & $50$ \\
$\lambda$ & \textit{First stage} -- Line 4 & $30$ & $10$ & $30$ & $30$ \\
$lr$ & \textit{First stage} -- Line 11 & $0.05$ & $0.05$ & $0.01$ & $0.05$ \\
$E_2$ & \textit{Second stage} -- Line 14 & $10$ & $10$ & $10$ & $5$ \\
$\lambda$ & \textit{Second stage} -- Line 17 & $10$ & $5$ & $10$ & $50$ \\
$lr$ & \textit{Second stage} -- Line 23 & $0.001$ & $0.001$ & $0.01$ & $0.1$ \\
$M$ & \textit{Requirements} & $5$ & $5$ & $5$ & $5$ \\
\bottomrule
\end{tabular}
\end{center}
    \caption{Hyperparameters used for each dataset  - \textbf{part 1}.}
    \label{tab:hyperparams}
\end{table*}

%% file: sec/suppl/hyperparams_ood.tex
\begin{table*}[t]
    \begin{center}
\rowcolors{2}{lightgray}{}
\begin{tabular}{lcccccc}
\toprule
\textbf{} & \textbf{Refs. to Algorithm} & \textbf{\shortspliteurosat} & \textbf{\shortsplitresisc} & \textbf{\shortsplitcropdiseases} & \textbf{\shortsplitisic} & \textbf{\shortsplitchestx} \\
\midrule
$E_1$ & \textit{First stage} -- Line 1 & $10$ & $10$ & $10$ & $50$ & $10$ \\
$\lambda$ & \textit{First stage} -- Line 4 & $30$ & $10$ & $30$ & $5$ & $30$ \\
$lr$ & \textit{First stage} -- Line 11 & $0.05$ & $0.05$ & $0.01$ & $0.01$ & $0.05$ \\
$E_2$ & \textit{Second stage} -- Line 14 & $10$ & $10$ & $10$ & $10$ & $10$ \\
$\lambda$ & \textit{Second stage} -- Line 17 & $5$ & $5$ & $5$ & $10$ & $5$ \\
$lr$ & \textit{Second stage} -- Line 23 & $0.1$ & $0.01$ & $0.001$ & $0.01$ & $0.05$ \\
$M$ & \textit{Requirements} & $5$ & $5$ & $5$ & $5$ & $5$ \\
\bottomrule
\end{tabular}
\end{center}
    \caption{Hyperparameters used for each dataset - \textbf{part 2}.}
    \label{tab:hyperparams_ood}
\end{table*}

%% file: tables/params.tex
\begin{table}
\centering
\rowcolors{2}{lightgray}{}
\begin{tabular}{cc}
\toprule
\textbf{Model} & \textbf{ \#params (millions)} \\
\midrule
\textbf{Joint, Fine-Tune, LwF} & $86$ \\
\textbf{GDumb, DER++, SLCA} & $86$ \\
\textbf{L2P} & $0.12$ \\
\textbf{DualPrompt} & $0.41$ \\
\textbf{CODA-Prompt} & $3.91$ \\
\textbf{AttriCLIP} & $0.0998$ \\
\textbf{PromptFusion} & $0.35$ \\
\textbf{STAR-Prompt} & $3.89$ \\
\bottomrule
\label{tab:number_of_gaussians}
\end{tabular}
    \caption{The number of learnable parameters for each tested method. Note that all these methods use a \backbone as the main classification architecture (see \cref{sec:experiments}).}
    \label{tab:tab_params_abl}
\end{table}

%% file: tables/result_abl_suppl.tex
\begin{table}[t]
\centering
\small
\rowcolors{2}{lightgray}{}
\resizebox{\textwidth}{!}{
\begin{tabular}{lccccc}
\midrule
\textbf{Model} & \textbf{\shortsplitcifar} & \textbf{\shortsplitcub}  & \textbf{\shortsplitresisc} & \textbf{\shortsplitcropdiseases} & \textbf{\shortsplitchestx} \\
\midrule
\textbf{\methnam}  & \ressmallb{90.12}{0.32}{4.58}    & \ressmallb{84.10}{0.28}{6.01}  & \ressmallb{92.28}{0.54}{4.58}  & \ressmallb{94.92}{0.60}{4.58}            & \ressmallb{41.85}{2.63}{4.58}            \\
\midrule
\multicolumn{6}{c}{\textbf{Ablations on two-level prompting}} \\
\midrule
\textit{Classify with first-level keys $\pmb{w}_c$}                   & \ressmall{83.49}{0.16}{5.77}                 & \ressmall{81.31}{0.20}{}  & \ressmall{90.80}{0.34}{}                 & \ressmall{88.36}{0.74}{}                 & \ressmall{31.70}{0.77}{}                 \\
\textit{w/o first-level prompts}               & \ressmall{87.67}{0.37}{}                 & \ressmall{81.34}{0.17}{}  & \ressmall{85.85}{0.69}{}                 & \ressmall{89.92}{1.76}{}                 & \ressmall{37.27}{3.96}{}                 \\
\midrule
\multicolumn{6}{c}{\textbf{Other secondary ablations}} \\
\midrule
\textit{Prefix Tuning (no residuals)}      & \ressmall{86.70}{0.59}{}  & \ressmall{82.67}{0.14}{} & \ressmall{84.26}{0.35}{}          & \ressmall{95.06}{0.34}{}               & \ressmall{39.28}{4.14}{} \\
\textit{w/o Generative Replay}             & \ressmall{88.72}{0.41}{}  & \ressmall{82.21}{0.39}{} & \ressmall{88.2}{0.79}{}           & \ressmall{88.82}{1.25}{}               & \ressmall{37.66}{2.12}{}                 \\
\textit{w.\ Unimodal Generative Replay}    & \ressmall{90.07}{0.34}{}  & \ressmall{83.16}{0.14}{} & \ressmall{92.29}{0.51}{}          & \ressmall{94.21}{0.44}{}               & \ressmall{38.77}{2.76}{}                 \\
\textit{w/o Confidence Modulation}         & \ressmall{89.92}{0.18}{}  & \ressmall{83.74}{0.30}{} & \ressmall{92.05}{0.44}{}          & \ressmall{93.97}{0.21}{}               & \ressmall{39.28}{3.17}{}                 \\
\midrule
\end{tabular}}
    \caption{Ablative studies on \methnam (Final Avg. Acc. ± std dev).}
    \label{tab:abl_suppl}
\end{table}

%% file: tables/n_gaussians_table.tex
\begin{table}[t]
\centering
\rowcolors{2}{lightgray}{}
\begin{tabular}{lc}
\toprule
\textbf{$M$}                              & \textbf{Final Average Accuracy} \\
\midrule
$2$ & $88.94$ \\
$5$ & $89.37$ \\
$10$ & $89.16$ \\
$20$ & $89.58$ \\
\bottomrule
\label{tab:number_of_params}
\end{tabular}
    \caption{Impact of the number of Gaussians.}
    \label{tab:tab_m_abl}
\end{table}

%% file: tables/abl_concat_vs_sum.tex
\begin{table*}
    \begin{center}
\rowcolors{2}{lightgray}{}
\begin{tabular}{lcccc}
\toprule
\textbf{Conditioning method} & \textbf{\shortsplitimagenet} & \textbf{\shortsplitcifar} & \textbf{\shortsplitcars} & \textbf{\shortsplitcub} \\
\midrule
\textbf{Prefix Tuning }(\textit{one token}) & $71.45$ & $85.90$ & $52.66$ & $80.12$ \\
\textbf{Semantic Residuals} &  \bf 89.16 & \bf 90.16 & \bf 86.50 &  \bf 85.24 \\
\bottomrule
\end{tabular}
\end{center}
    \caption{Comparing Semantic Residuals with Prefix Tuning. In the latter, we use a single-key and a single-value prompt token.}
    \label{tab:results_c_vs_s}
\end{table*}